\documentclass[twocolumn,journal]{IEEEtran}
\usepackage[T1]{fontenc}
\usepackage[latin9]{inputenc}
\usepackage{color}
\usepackage{array}
\usepackage{booktabs}
\usepackage{multirow}
\usepackage{amstext}
\usepackage{amsthm}
\usepackage{amssymb}
\usepackage{stackrel}
\usepackage{graphicx}
\usepackage[unicode=true,
 bookmarks=true,bookmarksnumbered=true,bookmarksopen=true,bookmarksopenlevel=1,
 breaklinks=false,pdfborder={0 0 0},pdfborderstyle={},backref=false,colorlinks=false]
 {hyperref}
\hypersetup{pdftitle={Your Title},
 pdfauthor={Your Name},
 pdfpagelayout=OneColumn, pdfnewwindow=true, pdfstartview=XYZ, plainpages=false}

\makeatletter

\newcommand{\noun}[1]{\textsc{#1}}
\providecommand{\tabularnewline}{\\}

\let\oldforeign@language\foreign@language
\DeclareRobustCommand{\foreign@language}[1]{%
  \lowercase{\oldforeign@language{#1}}}
\theoremstyle{plain}
\newtheorem{thm}{\protect\theoremname}
\theoremstyle{definition}
\newtheorem{defn}[thm]{\protect\definitionname}

\usepackage[caption=false,font=footnotesize]{subfig}

\@ifundefined{showcaptionsetup}{}{%
 \PassOptionsToPackage{caption=false}{subfig}}
\usepackage{subfig}
\makeatother

\providecommand{\definitionname}{Definition}
\providecommand{\theoremname}{Theorem}

\begin{document}
\newcommand{\sectionname}{Section}
\newcommand{\eqname}{Eq.}
\title{Lamarckian Platform: Pushing the Boundaries of Evolutionary Reinforcement
Learning towards Asynchronous Commercial Games}
\author{Hui~Bai, Ruimin~Shen, Yue~Lin, Botian Xu, and Ran~Cheng,~\IEEEmembership{Senior Member,~IEEE}\thanks{H. Bai and R. Shen contribute equally to this work.}\thanks{H. Bai, B. Xu, and R. Cheng are with the Guangdong Key Laboratory
of Brain-Inspired Intelligent Computation, Department of Computer
Science and Engineering, Southern University of Science and Technology,
Shenzhen 518055, China. E-mail: \protect\href{http://huibaimonky@163.com}{huibaimonky@163.com}, 
\protect\href{http://btx0424@outlook.com}{btx0424@outlook.com}, \protect\href{http://ranchengcn@gmail.com}{ranchengcn@gmail.com}. \emph{(Corresponding
author: Ran Cheng)}}\thanks{R. Shen and Y. Lin are with the NetEase Games AI Lab, Guangzhou 510653,
China. E-mail: \protect\href{http://shenruimin@corp.netease.com}{shenruimin@corp.netease.com},
\protect\href{http://gzlinyue@corp.netease.com}{gzlinyue@corp.netease.com}.} \thanks{This work was supported by the National Natural Science Foundation
of China (No. 61906081), the Shenzhen Science and Technology Program
(No. RCBS20200714114817264), the Guangdong Provincial Key Laboratory
(No. 2020B121201001), and the Program for Guangdong Introducing Innovative
and Entrepreneurial Teams (Grant No. 2017ZT07X386).}}
\markboth{IEEE TRANSACTIONS ON GAMES, VOL. , NO. , MONTH YEAR}{}
\maketitle
\begin{abstract}
Despite the emerging progress of integrating evolutionary computation
into reinforcement learning, the absence of a high-performance platform
endowing composability and massive parallelism causes non-trivial
difficulties for research and applications related to asynchronous
commercial games. Here we introduce Lamarckian\footnote{The code and demonstrational setup of Lamarckian are publicly available at \href{https://github.\%20com/lamarckian/lamarckian}{https://github. com/lamarckian/lamarckian}.}
\textendash{} an open-source platform featuring support for evolutionary reinforcement
learning scalable to distributed computing resources. To improve the
training speed and data efficiency, Lamarckian adopts optimized communication
methods and an asynchronous evolutionary reinforcement learning workflow.
To meet the demand for an asynchronous interface by commercial games
and various methods, Lamarckian tailors an asynchronous Markov Decision
Process interface and designs an object-oriented software architecture
with decoupled modules. In comparison with the state-of-the-art RLlib,
we empirically demonstrate the unique advantages of Lamarckian on
benchmark tests with up to 6000 CPU cores: i) both the sampling efficiency
and training speed are doubled when running PPO on Google football
game; ii) the training speed is 13 times faster when running PBT+PPO
on Pong game. Moreover, we also present two use cases: 
i) how Lamarckian is applied to generating behavior-diverse game AI;
ii) how Lamarckian is applied to game balancing tests for an asynchronous commercial game.

\end{abstract}


\section{Introduction}

Reinforcement learning (RL), as a powerful tool for
sequential decision-making, has achieved remarkable successes in a
number of challenging tasks varying from board games \cite{Silver2016},
arcade games \cite{Mnih2015}, robot control \cite{Lillicrap2015},
scheduling problems \cite{Zhang2020} to autonomous driving \cite{Kiran2021}.
Despite that RL algorithms have been widely assessed on game benchmarks (e.g., Atari games
\cite{Brockman2016}, ViZDoom \cite{Wydmuch2019}, and DeepMind Lab
\cite{Beattie2016}), the applications of RL in commercial games (e.g., StarCraft I \cite{Xu2019}
\& II \cite{Liu2021c}, and Dota2 \cite{Berner2019}) have raised new issues to be considered, e.g.,  partially observed maps, large state
space and action space, delayed credit assignment, etc.
Besides, when applying RL to real-world scenarios, there are also a number of technical challenges such as brittle convergence properties caused by sensitive hyperparameters, temporal credit assignment with long time horizons and sparse rewards, difficult credit assignments
in multi-agent reinforcement learning, lack of diverse exploration,
a set of conflicting objectives for rewards, etc.
% Consequently, the training costs of RL agents become increasingly expensive due to the challenges brought by these new features.
% For example, OpenAI trains a Dota2 playing agent with thousands of GPUs and CPUs over multiple months. 

To meet the above challenges, recently, there has been an emerging progress in integrating RL with evolutionary
computation (EC) to address the above challenges. 
In first-person
multiplayer games, the Population Based Training (PBT) trains a population
of agents to dynamically optimize hyperparameters for self-play \cite{Jaderberg2019}.
In various benchmark games of OpenAI Gym\textcolor{black}{{} \cite{Brockman2016},}
the Evolutionary Reinforcement Learning \cite{Khadka2018} and Collaborative
Evolutionary Reinforcement Learning \cite{Khadka2019} take the advantages
of EC to make up for some deficiencies in RL, such as difficult credit
assignment, lack of effective exploration, and brittle convergence
through a fitness metric.\textcolor{black}{{} EMOGI generates desirable
styles of game artificial intelligence (AI) by a multi-objective EC algorithm \cite{Shen2020a}.
Wuji combines EC and RL for automated game testing to detect game
bugs by exploring states as much as possible \cite{Zheng2019}. Besides,
under the names of neuroevolution/}derivative-free reinforcement learning,
Evolution strategies (ESs) and Genetic Algorithms (GAs) have been
used to optimize policy networks, which avoids the gradient vanishing
problem \cite{Qian2021}.\textcolor{black}{{} }In brief, the literature
has demonstrated promising potentials for integrating EC to RL, despite
that the development of related research and application is still
in its infancy.

While reinforcement learning enjoys the advances of several state-of-the-art
platforms or frameworks \cite{Castro2018,Hafner2017,Kuettler2019,Liang2018},
the literature is still in the absence of a platform or framework
specifically tailored to evolutionary reinforcement learning (EvoRL)\footnote{In this work, the general \emph{evolutionary reinforcement learning}
is abbreviated as \emph{EvoRL} for short.}. 
Moreover, current distributed platforms do not use computational
resources efficiently, and thus the training of RL is extremely time-consuming
for complex commercial games. Therefore, Lamarckian exactly meets
the rigid demand for such a highly decoupled, high-performance, scalable
implementation of a distributed architecture, to support research
and engineering in EvoRL. Most importantly, Lamarckian fully supports
the implementations of evolutionary multi-objective optimization,
which has demonstrated promising potentials in solving specific RL
problems involving multiple objectives to be considered simultaneously \cite{Vamplew2010,Abels2019}.  

Intrinsically, EvoRL can be seen as an evolutionary distributed RL paradigm requiring delicate management of computing resources for asynchronous training.
Hence, in addition to meeting the rigid demand of EvoRL as mentioned above, Lamarckian is also dedicated to improving data efficiency and training speed of general distributed RL.
Despite that some recent works have successfully scaled RL methods (e.g. PPO \cite{Schulman2017} and IMPALA \cite{Espeholt2018}) to large-scale distributed computing systems \cite{Berner2019},
the \emph{policy-lag} \cite{Espeholt2018} is still an open issue.
In distributed RL, policy-lag happens when a learner policy is several updates ahead of an actor's policy when an update occurs, thus causing severe defects in convergence and stability.
To address the issue of policy-lag, methods such as clipped surrogate objective~\cite{Schulman2017} and V-trace \cite{Espeholt2018} have been proposed. 
Apart from the delicate designs from methodology perspective, it has been evidenced that the improvements from the engineering
perspective could also significantly alleviate policy-lag, as did in the case of SEED RL ~\cite{Espeholt2020}.
%such as the effective utilization of TPUs by a centralized inference RL framework and a fast communication layer in SEED RL .
Specifically, Lamarckian is tailored for such a purpose from two aspects.
First, an asynchronous tree-shaped data
broadcasting method is proposed to reduce the policy-lag and alleviate
the communication bottleneck of learners.
Second, having inherited high scalability of the
state-of-the-art Ray \cite{Moritz2018}, Lamarckian further improves
the efficiency of distributed computing and increases throughput by
coupling Ray with ZeroMQ. 

%Moreover, apart from the decentralized inference RL methods such as PPO where a learner/actor are expected to execute model inference in a distributed system, other distributed RL methods methods can also benefit from the tailored communication methods.

From the engineering perspective, game environmental interfaces provide a system of API components allowing players to interact with the game story and break into the game space. For example, the interfaces receive actions from players and return the next state of the game environment to the players.
In this paper, game environmental interfaces can
be divided into two types: synchronous interfaces and asynchronous
interfaces. In synchronous interfaces (e.g., OpenAI Gym), a tick moves
to the next tick only when it receives actions from all players. In
asynchronous interfaces (e.g., most commercial games), each player
is controlled independently and asynchronously. Consequently, if a
player is off-line, the other players and the main tick will continue
moving without waiting. 
To meet such requirements of asynchronous commercial game environments, players with different roles are expected to interact with the environment independently.
Thus, Lamarckian is designed on the basis
of an asynchronous Markov Decision Process (MDP, e.g., modeled by
game playing) interface, which is highly decoupled by object-oriented
programming. With these tailored designs, Lamarckian is shipped with
a number of representative algorithms in EC and RL algorithms as summarized
in \tablename~\ref{tab:The-implemented-main-stream}. By simple
configurations, EC algorithms can be easily integrated with RL algorithms
in Lamarckian. Meanwhile, Lamarckian has good scalability in terms
of adding new algorithms, new environments, or new DNN models. In
brief, our main contributions of Lamarckian can be summarized as:
\begin{itemize}
\item \textcolor{black}{A highly decoupled, high-performance, scalable platform
tailored for EvoRL is delivered.}
\item \textcolor{black}{To accelerate training speed in large-scale distributed
computing, two methods are proposed from the engineering perspectives:
i) broadcasting data from the learner to actors by an asynchronous
tree-shaped data broadcasting method; and ii) coupling high scalability of Ray with high efficiency of ZeroMQ.}
\item \textcolor{black}{A highly decoupled, asynchronous MDP interface is
designed for asynchronous commercial game environments.}
\item \textcolor{black}{An object-oriented software architecture with decoupled
modules is developed to support various problems and algorithms in
RL and EC.}
%\item \textcolor{black}{Lamarckian is assessed by large-scale benchmarks and applied to a real-world }asynchronous\textcolor{black}{{} commercial game. }
\end{itemize}

The organization of the paper is as follows. \sectionname~\ref{sec:Background-and-Related}
gives the background and related work. \sectionname~\ref{sec:Asynchronous Distributed Designs} describes the asynchronous distributed designs. \sectionname~\ref{sec:Benchmark Experiments} provides the benchmark experiments, and \sectionname~\ref{sec:Use Cases} provides two use cases. Finally, \sectionname~\ref{sec:Conclusion} concludes
the paper.

\begin{table}[tbh]
\caption{\label{tab:The-implemented-main-stream}The main-stream algorithms
and benchmarks implemented in Lamarckian.}

\hfill{}\scriptsize%
\begin{tabular}{lc}
\toprule 
Modules & Implementations\tabularnewline
\midrule
RL algorithms & A3C \cite{Mnih2016}, PPO \cite{Schulman2017}, IMPALA \cite{Espeholt2018},
and DQN \cite{Mnih2013}, etc.\tabularnewline
EvoRL algorithms & PBT+self-play \cite{Jaderberg2019}, EMOGI \cite{Shen2020a}, Wuji
\cite{Zheng2019}, etc.\tabularnewline
RL benchmarks & OpenAI Gym \cite{Brockman2016}, Google football \cite{Kurach2020},
etc.\tabularnewline
EC algorithms & single-objective GA \cite{Ishibuchi2006a}, multi-objective NSGA-II
\cite{Deb2002a}, etc.\tabularnewline
EC benchmarks & multi-objective DTLZ \cite{Deb2005b} and ZDT \cite{Zitzler2000},
etc.\tabularnewline
\bottomrule
\end{tabular}\hfill{}
\end{table}

\section{{Background and }Related Work}\label{sec:Background-and-Related}

In this section, we first present the definition
of MDP and the formulation of RL in \sectionname~\ref{subsec:MDP-and-RL},
and then describe EC and EvoRL in \sectionname~\ref{subsec:EC-and-EvoRL}.
Next, we discuss the previous RL platforms and frameworks in \sectionname~\ref{subsec:RL-Systems-and}.
Finally, we summarise and discuss essential terminologies in \sectionname~\ref{subsec:discussion of terms}.

\subsection{{RL and MDP\label{subsec:MDP-and-RL}}}

{RL is a special branch of AI considering the interactions
between agents and an environment towards maximum rewards. The problem
that an agent acts in a stochastic environment by sequentially choosing
actions over a sequence of time steps to maximise a cumulated reward
can be modeled as a Markov Decision Process. }
\begin{defn}
{(Markov Decision Process) An MDP is usually defined
as $<S,A,T,R,\rho_{0},\gamma>$, with a state space $S$, an action
space $A$, a stochastic transition function $T$: $S\times A\rightarrow P(S)$
representing the probability distribution over possible next states,
a reward function $R$: $S\times A\rightarrow\mathbb{R}$, an initial
state distribution $\rho_{0}$: $S\rightarrow\mathbb{R}_{\in[0,1]}$,
and a discount factor $\gamma\in[0,1)$.}

At each discrete time step $t$, given the current state $s_{t}\in S$,
the agent selects actions $a_{t}\in A$ according to its policy {$\pi_{\theta}:S\rightarrow P(A)$,
where $P(A)$ is the set of probability measures on $A$ and $\theta\in\mathbb{R}^{n}$
is a vector of $n$ parameters, and $\pi_{\theta}(a_{t}|s_{t})$ is
the conditional probability density at $a_{t}$ given the input $s_{t}$
associated with the policy. The agent's objective is to learn a policy
to maximize the expected cumulative discounted reward from the start
state:}

{
\begin{equation}
J(\pi)=\mathbb{E}_{\rho_{0},\pi,T}\left[\stackrel[t=0]{\infty}{\sum}\gamma^{t}r_{t}\right],
\end{equation}
where $s_{0}\sim\rho_{0}(s_{0})$, $a_{t}\sim\pi(s_{t})$, $s_{t+1}\sim T(\cdot|s_{t},a_{t})$,
and $r_{t}=R(s_{t},a_{t})$.}

In practice, asynchronous commercial games often involve multiple players, where the main tick will continue moving without waiting off-line players.
In this case, an asynchronous  MDP interface will be required for dealing with the interactions between agents and
the environment. 
%To this end, Lamarckian tailors an asynchronous MDP interface, where agents are decoupled according to their requirements (e.g., whether they need return reward or not).
\end{defn}

\subsection{{EC and EvoRL\label{subsec:EC-and-EvoRL}}}

Evolutionary computation (EC) generally refers to the family of population-based stochastic optimization
algorithms (e.g., Population Based Training (PBT) \cite{Jaderberg2017},
Evolutionary Strategy (ES) \cite{Silver2016}, Genetic Algorithm (GA)
\cite{Whitley1994}, etc.) inspired by natural evolution. 
%Inspired by natural evolution, an EC algorithm is expected to search towards optima by iteratively performing operations such as variation (e.g., crossover and mutation), evaluation, and selection. 
Specifically, an EC algorithm first initializes a population
of candidate solutions, and then it enters an iterative loop: 
the candidate solutions in the current population are pairwisely mated to undergo the \emph{permutation operations} (i.e., crossover and mutation)
to generate new offspring candidate solutions; the offspring candidate solutions
are \emph{evaluated} by task-related performance indicators to obtain their
fitness values (a.k.a. objective values); the offspring candidate solutions are merged with the ones in the current population to be selected for the next iteration (a.k.a generation).
After a number of generations as above, ideally, an EC algorithm will end up with a population of candidate solutions approximating the global optima of the optimization problems as given.

The optimization problems in RL tasks often involve complex characteristics, while EC has been found to be a powerful tool for dealing with them~\cite{Qian2021}.
On the one hand, EC algorithms require no gradient information and is widely applicable to problems
without explicit objective functions by quality diversity (QD) \cite{Pugh2016}
or novelty search (NS) \cite{Lehman2011a}.
On the other hand, thanks to the population-based nature, EC algorithms are inherently robust to dynamic changes that widely exist
in real-world applications of RL (e.g., sim-to-real transfer in robot
control \cite{Zhao2020}). 
%In contrast, the gradient-based methods for RL may lose their power for easily falling into the local optima or become ineffective without gradient information.

As an emerging research direction of RL, EvoRL is
exactly dedicated to meeting the challenges of various key research problems
in RL research by dealing the complex optimization problems as involved, including but not limited to policy search \cite{Pourchot2018},
reward shaping \cite{Liu2019a}, exploration \cite{Khadka2018}, hyperparameter
optimization \cite{Jaderberg2019}, meta-RL \cite{Co-Reyes2021},
multi-objective RL \cite{Yang2019}, etc.

\begin{table}[tbh]
\caption{\label{tab:Compare-Lamarckian-and}Summary of Lamarckian and three
representative distributed reinforcement learning frameworks or platforms.}

\hfill{}\scriptsize%
\begin{tabular}{lcccc}
\toprule 
Frameworks & Single & Multi- & Tailored EvoRL & Asynchronous\tabularnewline
/Platforms & Agent & Agent & Framework/Workflow & MDP Interface\tabularnewline
\midrule
TLeague \cite{Sun2020} & $\checkmark$ & $\checkmark$ & $\times$ & $\times$\tabularnewline
Acme \cite{Hoffman2020} & $\checkmark$ & $\checkmark$ & $\times$ & $\times$\tabularnewline
RLlib \cite{Liang2018} & $\checkmark$ & $\checkmark$ & $\times$ & $\times$\tabularnewline
Lamarckian & $\checkmark$ & $\checkmark$ & $\checkmark$ & $\checkmark$\tabularnewline
\bottomrule
\end{tabular}\hfill{}
\end{table}

\subsection{RL Platforms and Frameworks\label{subsec:RL-Systems-and}}

Since an RL agent is trained on a large number of samples generated by interacting with its environment, the training speed is highly related to the sampling efficiency. Hence, the sampling efficiency is an important indicator to measure RL frameworks or platforms \cite{Liang2018,Hoffman2020}. This motivates the distributed framework to sample in parallel, where each of the multiple actors interacts with its environment independently. Except for the sampling efficiency, the sample staleness reflecting the policy-lag is another essential indicator \cite{Berner2019}. To keep learners and actors as consistent as possible, various synchronous methods are employed in state-of-the-art distributed frameworks and platforms. Specially, RLlib uses the synchronous data broadcasting where the learner stops policy optimization when sending data to actors \cite{Liang2018}. Acme applies the data storage system Reverb and its rate limiter to control the policy-lag, where the rate limiter will block faster processes until slower processes catch up \cite{Hoffman2020}. However, since 
both RLlib and Acme may have blocking or waiting processes for synchronization,
they are low-efficient in large-scale scenarios.

From the engineering perspective, most existing reinforcement learning platforms or frameworks tend
to scale at long-running program replicas for distributed execution
\cite{Castro2018,Hafner2017,Kuettler2019,Duan2016,Hoffman2020}, thus
causing difficulties in generalizing to complex architectures. In
contrast, RLlib scales well at short-running tasks by a Ray-based
hierarchical control model \cite{Liang2018}. Despite the high scalability,
however, RLlib suffers slow training efficiencies on large-scale distributed
computing environments, and therefore RLlib cannot generalize well to complex AI systems requiring
highly parallel data transmission between the learner and actors,
such as OpenAI Five \cite{Berner2019}. 

Among other state-of-the-art frameworks and platforms, SEED RL \cite{Espeholt2020}
and TorchBeast \cite{Kuettler2019} are two high-performance scalable
implementations of IMPALA \cite{Espeholt2018}. SURREAL \cite{Fan2018}
focuses on continuous control agents in robot manipulation tasks by
a distributed training framework. The recently proposed frameworks,
e.g., Arena \cite{Song2020}, TLeague \cite{Sun2020}, and MALib \cite{Zhou2021},
target at multi-agent reinforcement learning.

Despite the various platforms or frameworks as introduced above, 
none of them is highly parallel or tailored for evolutionary reinforcement learning, particularly,
facing the applications of asynchronous commercial games.  \textcolor{black}{\tablename~\ref{tab:Compare-Lamarckian-and}
summarizes the main features of Lamarckian, in comparison with several
representative RL platforms or frameworks from four aspects}.

\subsection{Discussions\label{subsec:discussion of terms}}
%To capture the concepts of essential terminologies, we summarise and provide comparative discussions. 
%For further clarifications, we will provide some discussions on some core terminologies as mentioned above.

%From the perspective of system designs for a platform, \emph{composability} is well-known as the degree of decoupling and reusability between different modules. 
% \emph{Synchronous} system has a blocking architecture, where one operation is being performed while other operations' instructions are blocked.

From the perspective of engineering designs, an \emph{asynchronous} system has a non-blocking architecture where multiple operations can run 
concurrently without waiting others to complete; 
a \emph{distributed} system is a computing environment where independent components run on different machines to achieve a common goal. 
Hence, \emph{distributed RL} and \emph{asynchronous RL} refer to engineering implementations of RL in the \emph{system level}.

On the contrary, \emph{EvoRL} aims to adopt EC methods to deal with the challenging issues for RL in the \emph{methodology level}.
Nonetheless, considering the population-based property of EvoRL, delicate \emph{asynchronous} system designs are particularly important for the high performance of EvoRL.

%The components of a asynchronous RL system can include parallel methods, sampling methods, communication methods, interaction methods, etc.
% Undoubtedly, algorithms implemented on an asynchronous/distributed RL system has better efficiency.
Therefore, we are motivated to improve efficiency for EvoRL as well as conventional RL by adopting a series of asynchronous system designs, including: 
\emph{asynchronous distributed EvoRL workflow} allowing several operations of EvoRL to execute in different machines concurrently without communication or information exchange, 
\emph{asynchronous sampling} allowing a learner to continue learning when sending its policy to actors;
\emph{asynchronous data broadcasting} allowing the simultaneous data transmission and reception mode based on asynchronous sampling, and \emph{asynchronous MDP interfaces} compatible with asynchronous commercial games.

\section{Asynchronous Distributed Designs}\label{sec:Asynchronous Distributed Designs}

To provide highly parallel and scalable implementations of distributed
evolutionary reinforcement learning algorithms, Lamarckian is designed
to be asynchronous distributed by considering four main aspects: the
distributed EvoRL workflow in {\sectionname~\ref{subsec:Distributed-EvoRL-Workflow}},
the acceleration of distributed computing in {\sectionname~\ref{subsec:Acceleration-of-Distributed}},
the asynchronous MDP interface in {\sectionname~\ref{subsec:Asynchronous-MDP-Interface}},
and the object-oriented software architecture in {\sectionname~\ref{subsec:Object-Oriented-Software-Archite}}.

\subsection{\textcolor{black}{Distributed EvoRL Workflow\label{subsec:Distributed-EvoRL-Workflow}}}

\begin{figure}[tbh]
\hfill{}\includegraphics[scale=0.4]{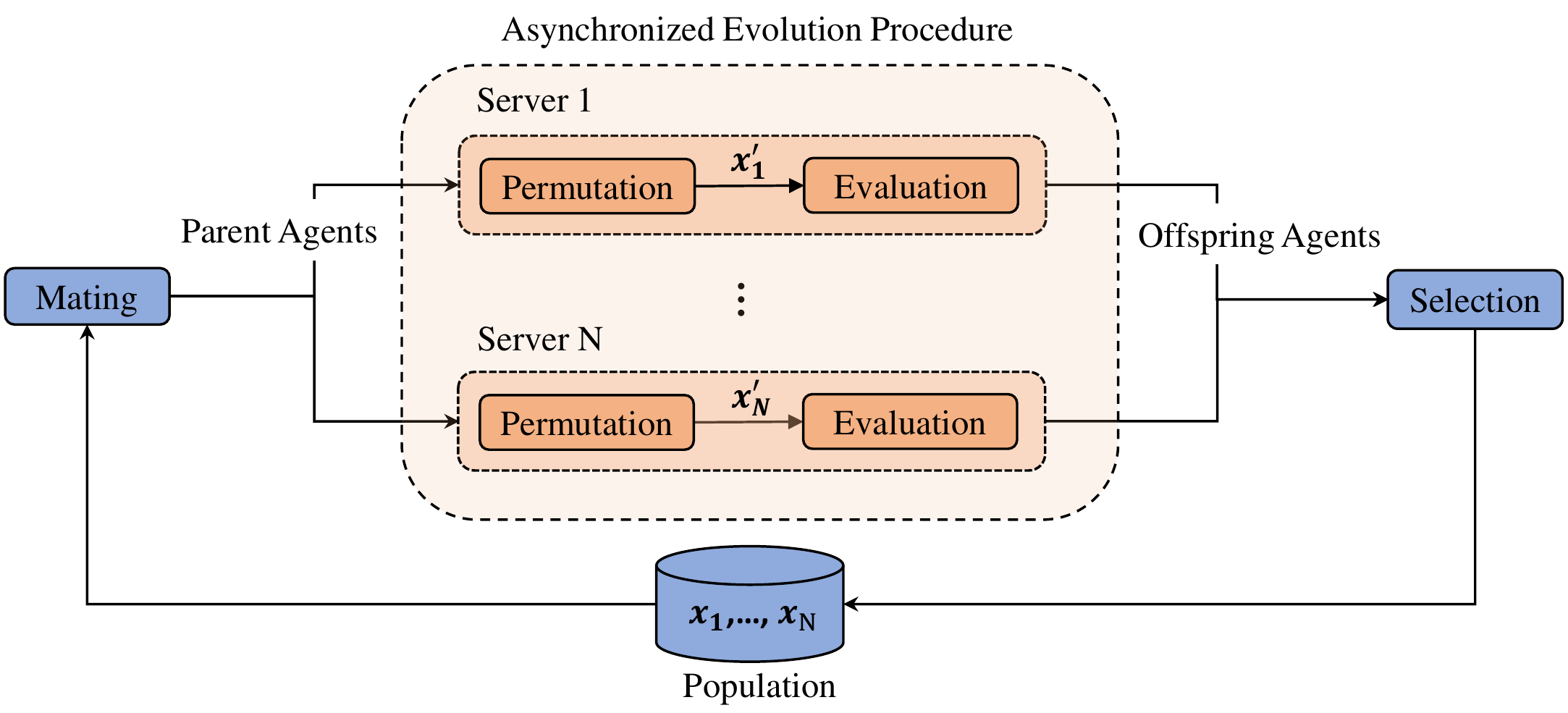}\hfill{}

\caption{\label{fig:Distributed Evolutionary Reinforcement Learning}Distributed
workflow of EvoRL. In the \emph{Asynchronous Evolution Procedure},
each pair of candidate solutions (a.k.a parents) are distributed in
an independent server to conduct permutation and evaluation in an
asynchronous manner.}
\end{figure}

Existing RL platforms merely support very limited single-objective
EC algorithms such as PBT. However, On the one hand, there are rich scenarios
involving multiple (instead of single) objectives to be optimized
simultaneously in RL tasks; on the other hand, EvoRL requires tailored
workflow design for general high-performance implementations.

\figurename~\ref{fig:Distributed Evolutionary Reinforcement Learning}
is the overview of the proposed distributed workflow for EvoRL: first,
a population of candidate solutions ($\mathbf{x}_{1}$, $\mathbf{x}_{2}$,
..., $\mathbf{x}_{N}$) is initialized; then, each candidate solution
enters the mating process to generate parent pairs; afterward\noun{,
}permutation operations are performed on each pair of mated candidate
solutions to generate offspring candidate solution; finally, the offspring
candidate solution is evaluated by an evaluator encapsulating configurable
RL algorithms or other task-related performance indicators.

Considering computational efficiencies, the evolution workflow is
asynchronous on each step, i.e., permutation, training, and evaluation.
Particularly, after evaluation, each offspring candidate solution
is sent to the collector server independently. Once the expected number
of offspring candidates solutions are obtained, a synchronized selection
will be performed to have the promising candidate solutions survive
to the next generation.

\subsection{Acceleration of Distributed Computing\label{subsec:Acceleration-of-Distributed}}

On top of the proposed distributed workflow, we further improve the
computing efficiency from two engineering perspectives.

\begin{figure}[tbh]
\hfill{}\includegraphics[scale=0.22]{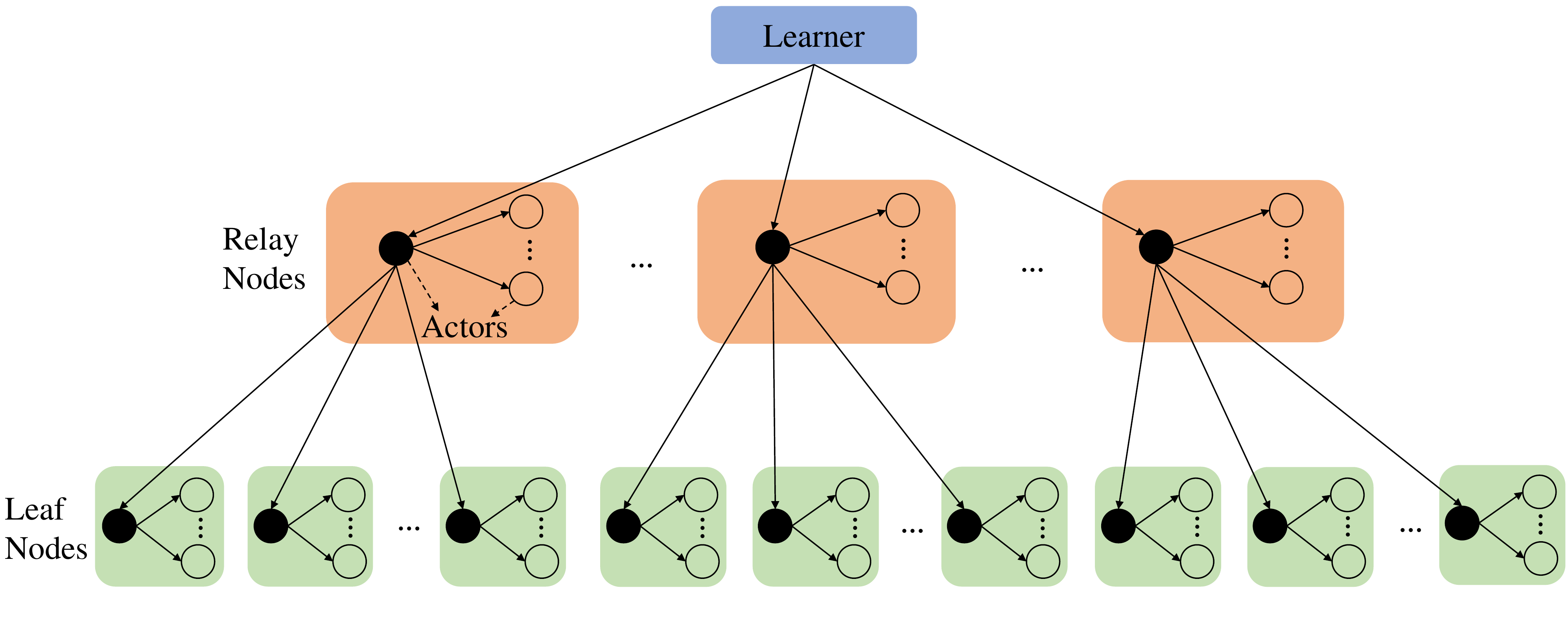}\hfill{}

\caption{\label{fig:Asynchronous Data Broadcasting} Asynchronous tree-shaped
data broadcasting. First, all actors are grouped by the machines where
they are located. In the same machine (i.e. each box), actors fetch
data by high-performance communication. Then, the learner sends data
to relay nodes. Finally, each relay node sends data to its connected
leaf nodes.}
\end{figure}

\begin{figure*}
\hfill{}
\subfloat[\label{fig:Code rewriting by Ray}Code rewriting by Ray.]{\includegraphics[scale=0.36]{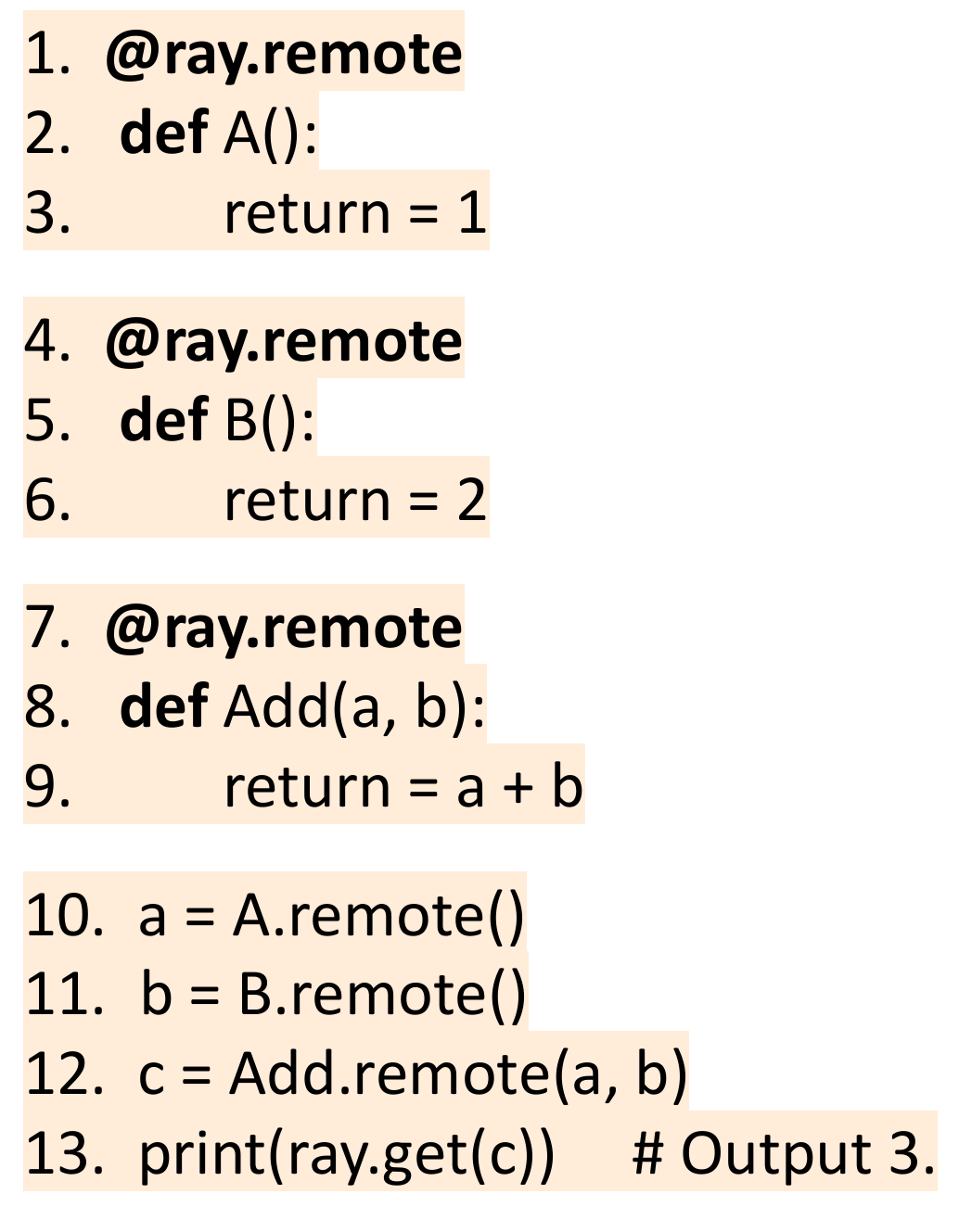}

}\hfill{}\subfloat[\label{fig:A Ray cluster}A Ray cluster.]{\includegraphics[scale=0.49]{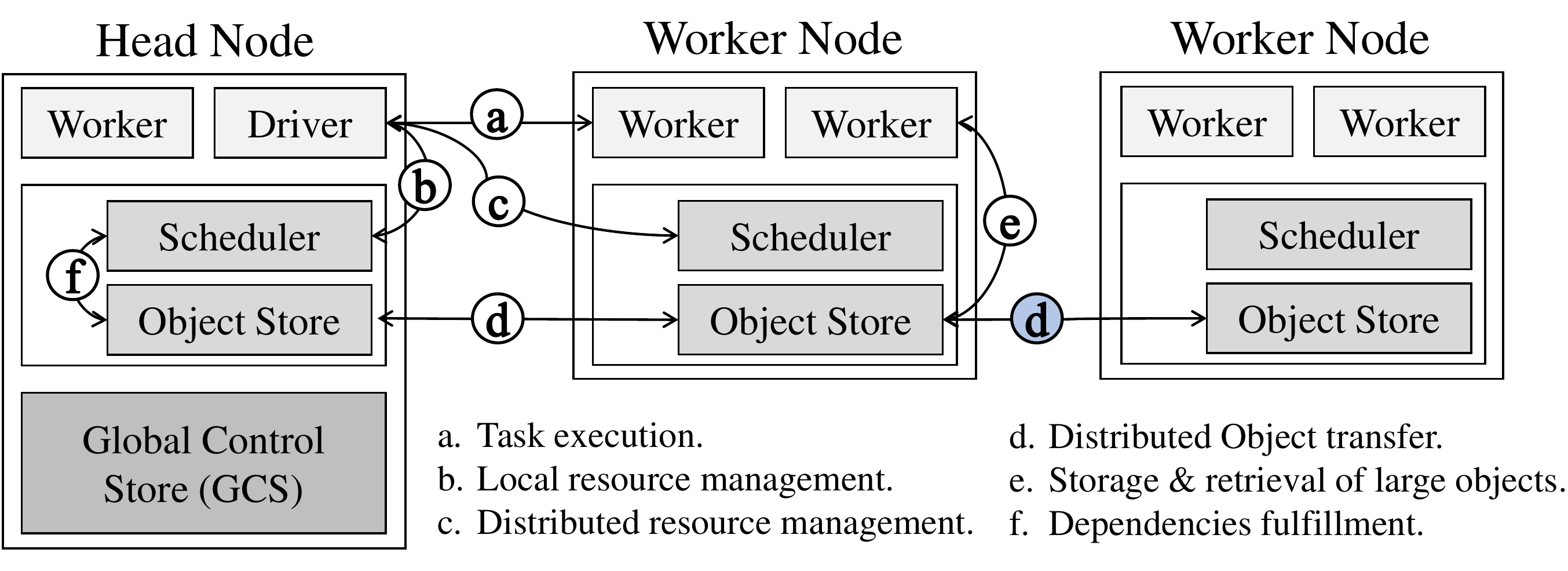}
}\hfill{}

\caption{(a) is an example of
using Ray to rewrite the code that outputs the result of $a+b$. (b)
is an example of a Ray cluster. The cluster architecture consists
of a head node and two worker nodes, where the head node has a global
control store (GCS) managing the system metadata, and each node has
its local worker, scheduler, object store. The protocols between these
components are presented in ``a'' to ``f'' in circles.}
\end{figure*}

\subsubsection{\label{subsec:Asynchronous-Tree-shaped-Data}Asynchronous Tree-shaped
Data Broadcasting}

It is common in an RL algorithm to require estimating a term in the
form of $\mathrm{E}_{\tau\sim\pi}\left[f(\tau)\right]$, where each
$\tau=(s_{0},a_{0},s_{1},a_{1},\dots,s_{T})$ is a trajectory generated
by carrying out the policy $\pi$. Since the expectation is taken
under the distribution induced by $\tau$, it is expected for on-policy
RL algorithms that the samples used for training should be as up-to-date
as possible to give an unbiased estimate.

For large-scale on-policy RL, when a learner is associated with a
large number of actors, a learner policy is potentially several updates
ahead of an actor's policy when an update occurs, thus causing the
policy-lag phenomenon. Since shorter data broadcasting time will directly
lead to a smaller policy-lag for asynchronous broadcasting, we propose
a tree-shaped data broadcasting method shown in \figurename~\ref{fig:Asynchronous Data Broadcasting}.

{The tree-shaped architecture is built automatically
and implicitly by getting available computational resources. The architecture
is a Complete N-ary Tree with no more than three layers for nodes
of any scale. Hence, }the out-degree of each node is approximated
by $\sqrt{n}$ ($n$ is the number of nodes). Each node represents
a machine with multiple CPU cores and/or GPUs. {The
GPU is given priority to the learner for a large amount of model inference.}
In \figurename~\ref{fig:Asynchronous Data Broadcasting}, we describe
the process of data broadcasting in three steps: first, all actors
are grouped by the machines where they are located, and in the same
machine, actors fetch data by high-performance communication (e.g.,
Inter-Process Communication or Shared Memory); then, the learner sends
data to relay nodes; finally, each relay node sends data to its connected
leaf nodes. 

With a flat layout, directly broadcasting to $n$ nodes would incur
$O(n)$ traffic, and the head node (usually the learner) becomes a
bandwidth bottleneck. In the worst case, it suffers an $O(n)$ delay
until the data to be received by all nodes. But with a tree-shaped
structure, it faces only $O(\sqrt{n})$ traffic, and the second-layer
relay nodes can operate in parallel. Consequently, this results in
a much smaller $O(\sqrt{n})$ delay. The effectiveness of this improvement
is demonstrated in the experiment section.

\subsubsection{Ray plus ZeroMQ}

Ray is a distributed computing framework for the
easy scaling of Python programs, which provides a simple and universal
API and only requires minimum modification to build distributed applications.
An example of using Ray to rewrite the code that outputs the result
of $a+b$ is provided in \figurename~\ref{fig:Code rewriting by Ray}, where the functions execute
as parallel and remote tasks by using the \emph{ray.remote} primitive.
The tasks can distribute in different nodes in a Ray cluster as in
\figurename~\ref{fig:A Ray cluster}, where the \emph{Add()} task executes in the head node, and
the \emph{A() }and\emph{ B()} tasks execute in the worker nodes, respectively.
The diagram illustrates an example of the cluster architecture and
protocols in brief. The head node has a global control store (GCS)
managing the system metadata. Each node has its local worker, scheduler, and object store. The protocols between these components, as presented
in ``a'' to ``f'' in the subfigure, are mostly over the Remote Procedure
Call (RPC) framework of gRPC. The details of Ray can refer
to the public \href{https://docs.google.com/document/d/1lAy0Owi-vPz2jEqBSaHNQcy2IBSDEHyXNOQZlGuj93c/preview\#}{document}.

Since the HTTP/2-based gRPC has large overhead, latency and complexity
of request/response chain in large-scale scenarios, the communication
efficiency is quite limited. In contrast, ZeroMQ is a powerful messaging
library featuring high-throughput and low latency. Moreover, it is
recognized that ZeroMQ is faster and more stable than HTTP/2 in communication.
Thus ZeroMQ is employed as transports in Ray to construct highly
efficient distributed computing framework for Lamarckian. Ray is used
for creating and managing remote objects (e.g., workers that collect
trajectories for training), and spawning processes correspondingly. The
underlying communication is through ZeroMQ sockets that implement
various messaging patterns, for example, Publish-Subscription for
broadcasting model weights and Push-Pull for distributing rollout
tasks. Benefiting from ZeroMQ, the large object transfer between worker nodes
is enabled to support data transfer between relay nodes and leaf nodes
in the tree-shaped data broadcasting, as demonstrated by ``d'' in  \figurename~\ref{fig:A Ray cluster}.

\subsection{Asynchronous MDP Interface\label{subsec:Asynchronous-MDP-Interface}}

To achieve high performance, flexible assembly, and scalability, Lamarckian
adopts the asynchronous MDP interface with highly decoupled properties.

\begin{figure}[tbh]
\subfloat[\label{fig:Synchronous-MDP-interface}Synchronous MDP interface of
Gym.]{\includegraphics[scale=0.32]{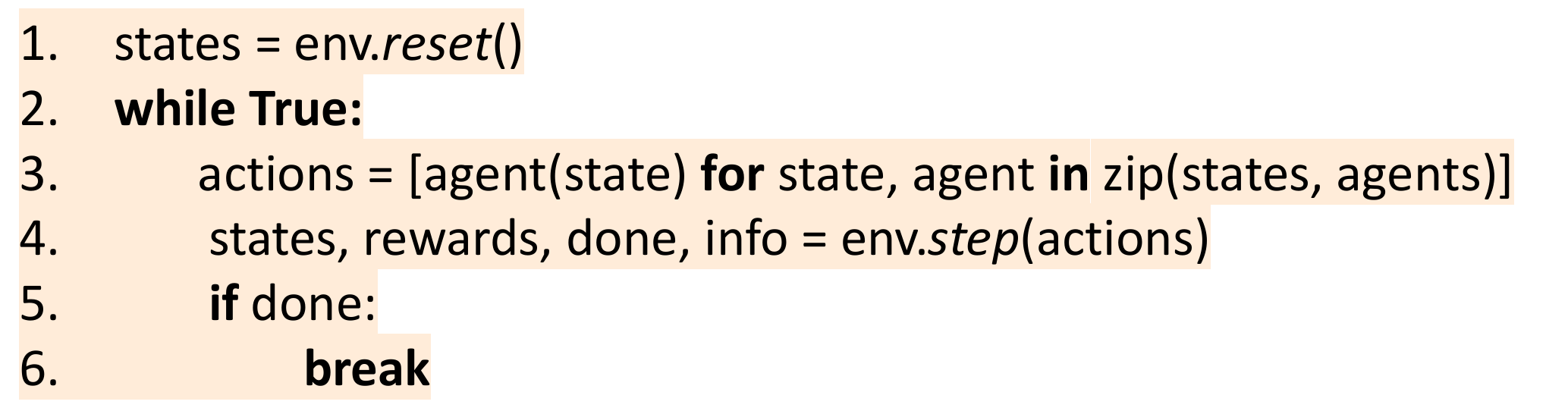}

}

\subfloat[\label{fig:Asynchronous-MDP-interface}Asynchronous MDP interface
of Lamarckian.]{\includegraphics[scale=0.32]{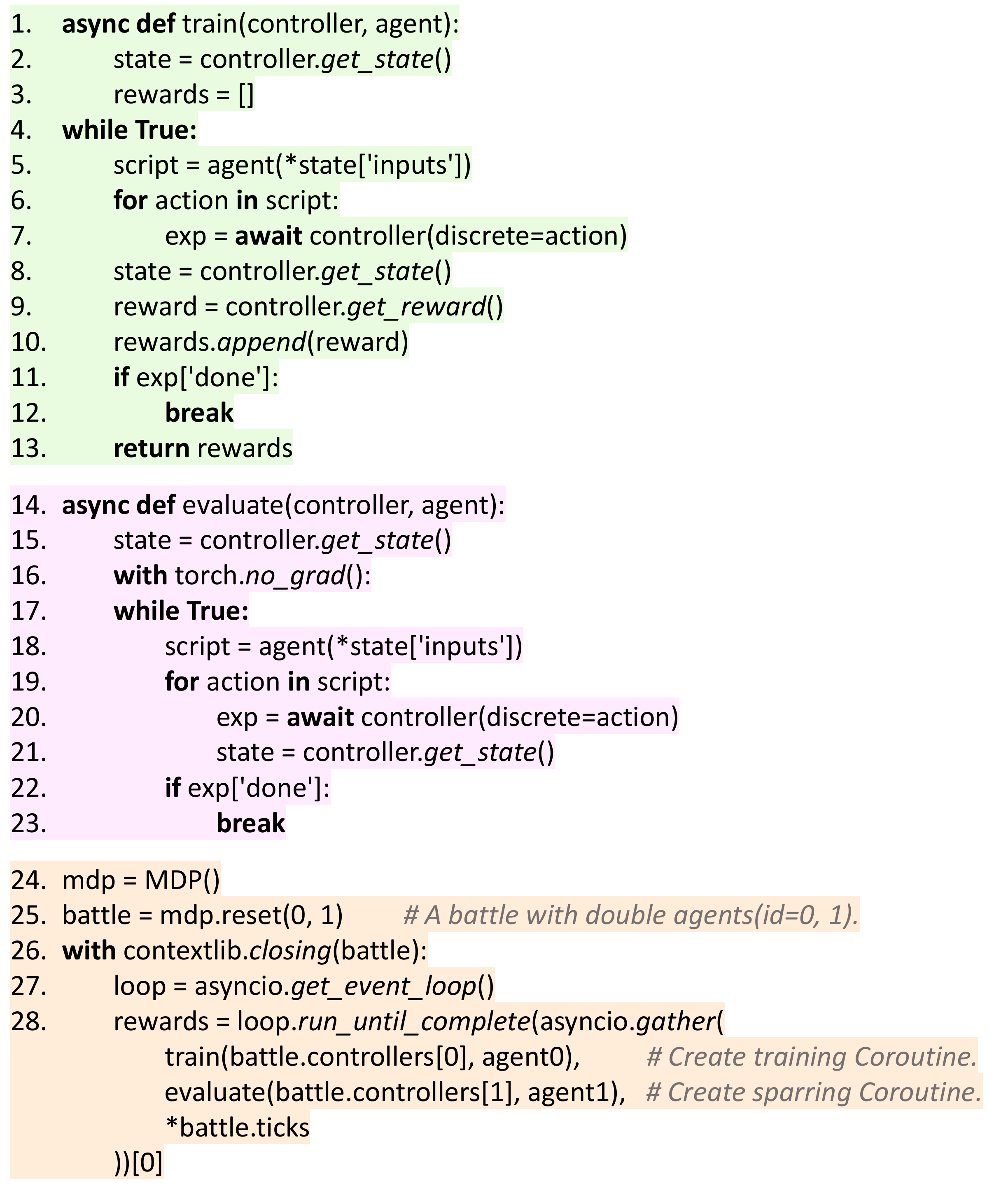}

}\caption{During the interaction between the environment and agents: the synchronous
MDP interface (a) is highly coupled, thus having poor compatibility
with asynchronous commercial games; in contrast, the asynchronous
MDP interface (b) decouples the agents with different functions, using
Coroutine-based controllers to control agents asynchronously.}
\end{figure}

\begin{figure*}[tbh]
\hfill{}\includegraphics[scale=0.6]{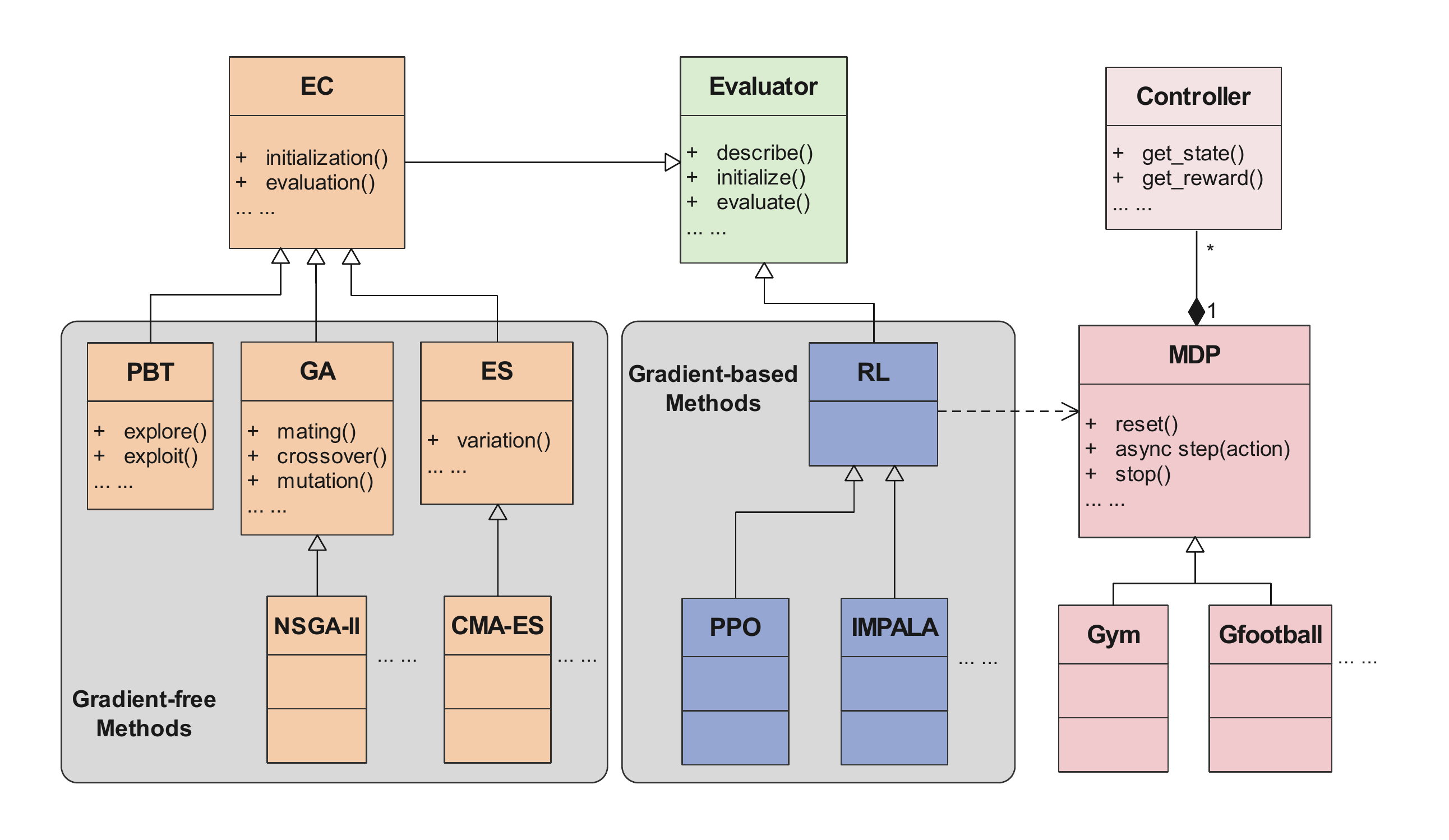}\hfill{}

\caption{\label{fig:Architecture of Lamarckian} The object-oriented software
architecture of Lamarckian: \textbf{EC} module provides a general
interface for EC algorithms; \textbf{Evaluator} module provides an
unified abstract interface for \textbf{EC} module and \textbf{RL}
module; \textbf{MDP} module provides an asynchronous interface that
decouples the agents with different functions, using Coroutine-based
controllers to control agents asynchronously.}
\end{figure*}

Most existing MDP interfaces are based on the procedure-oriented OpenAI
Gym, where a code example is given in (\figurename~\ref{fig:Synchronous-MDP-interface}).
The Gym-based MDP interfaces are synchronous in terms of two aspects:
1) the environment requires all generated actions from all agents
before it moves a step (line 3); 2) the environment simultaneously
returns states, rewards, and other information of all agents until
the agents perform all actions (line 4). Despite their simple implementations,
such Gym-based MDP interfaces suffer from three main limitations: 
\begin{itemize}
\item poor compatibility with asynchronous commercial game environments;
\item difficulties in implementing some specific RL techniques such as data
skipping technique \cite{Oh2021a} and scripted actions \cite{Berner2019}; 
\item computational redundancies caused by close couplings.
\end{itemize}
To address the above limitations, we have designed an object-oriented,
highly decoupled and asynchronous MDP interface (\figurename~\ref{fig:Asynchronous-MDP-interface}).
The proposed MDP interface is somehow similar to OpenAI Gym, but it
discriminates training agents and sparring agents by decoupling (line
28), where the training agent and the sparring agent can be respectively
implemented in two functions (i.e., the $train$ function and the
$evaluate$ function) according to their requirements (e.g., whether
they need to return rewards or not). One can define multiple controllers
for an MDP, where each controller controls an agent by a Coroutine
asynchronously. In contrast with Thread, Coroutine is more efficient
since it involves no scheduling overhead or synchronization overhead
on the level of operating systems.

Meanwhile, the independence and asynchrony of the proposed MDP interface
enable easy implementations of some specific RL techniques such as
scripted actions (lines 5-7 and 18-20). Moreover, the proposed MDP
interface well supports multiple inputs of states (lines 5 and 18),
multiple types of actions (e.g., discrete or continuous actions, lines
7 and 20), and multi-head value estimation \cite{Ye2020}. Consequently,
each agent can independently focus on its programming and execution
without considering others.

We demonstrate how to design an asynchronous MDP
by users. First, users should determine the types of agents according
to their actions and returned information. Second, each type of agent
is implemented in a separate function using Python Coroutine Syntax
(e.g., async; await; asyncio.get\_event\_loop; asyncio.gather). Specially,
as exemplified by the Gym codes in Fig. 4a, lines 3 and 4 are extended to line 2-10
in the \emph{train} function or
line 15-21 in the \emph{evaluate} 
function. Third, in the \emph{main}
function (line 28), a training agent and a sparring agent are initialized
and executed asynchronously, and then only the training agent returns
the rewards.

\subsection{Object-Oriented Software Architecture\label{subsec:Object-Oriented-Software-Archite}}

As shown in \figurename~\ref{fig:Architecture of Lamarckian} by
UML \cite{Medvidovic2002}, Lamarckian is designed on the basis of
an object-oriented software architecture comprehensively covering
key modules involved in EvoRL. In the subsection, we will introduce
each module one by one.

\textbf{Evaluator} module provides an unified abstract interface
for \textbf{EC} module and \textbf{RL} module, which includes three
general methods: $describe()$, $initialize()$, and $evaluate()$.
Specifically, $describe()$ returns the coding of a candidate solution
(e.g., integer coding, real coding, and neural network coding); $initialize()$
initializes a candidate solution with specific coding schemes; $evaluate()$
evaluates the performance of a candidate solution on specific objective/cost
function(s). Essentially, \textbf{Evaluator} provides a bridge between
\textbf{EC} module and \textbf{RL} module such that gradient-free
EC algorithms and gradient-based RL algorithms can work collaboratively
in RL tasks. 

\textbf{EC} module includes two general methods: $initialization()$
and $evaluation()$, which provides a general interface for EC algorithms
(e.g., PBT, GA, ES, etc.). Specifically, $initialization()$ initializes
a population of candidate solutions with corresponding coded decision
variables\footnote{In the context of RL, a \emph{decision variable} generally refers
to a \emph{weight} or \emph{hyperparameter}.}; $evaluation()$ evaluates the performance of a population of candidate
solutions by an instantiated evaluator.

\textbf{PBT} module includes two general methods $explore()$ and
$exploit()$. Implementation of any PBT-like algorithm falls into
this module.

\textbf{GA} module includes three general methods $mating()$, $crossover()$
and $mutation()$. Implementation of any GA based algorithm (e.g.,
NSGA-II \cite{Deb2002a}) falls into this module.

\textbf{ES} module includes a general method $variation()$. Implementation
of any ES based algorithm (e.g., CMA-ES \cite{Hansen2003}) falls
into this module.

{\textbf{RL} module provides a general interface
and primitives for state-of-the-art RL algorithms, such as A3C, PPO
and DQN.}

{\textbf{MDP} module provides an asynchronous interface.
Each MDP can include multiple controllers, where each controller controls
an agent. Any game environment can adapt to this asynchronous MDP
interface by modifications.}

Lamarckian can independently support EC algorithms by providing shared modules (e.g., GA and ES) and functions (e.g., initialization, evaluation, crossover and mutation). Thus, users can easily implement their EC algorithms by inheriting and reusing the existing modules and functions. To efficiently test the accuracy of implementing EC algorithms, users can also use the existing numerical optimization problems or implement their own problems by inheriting the Evaluator module. In addition, benefiting from the decoupled and object-oriented software architecture of the platform, users can flexibly apply different EC algorithms to different tasks by configuration commands or files.

\section{Benchmark Experiments}\label{sec:Benchmark Experiments}

\begin{figure}[tbh]
\hfill{}\includegraphics[scale=0.2]{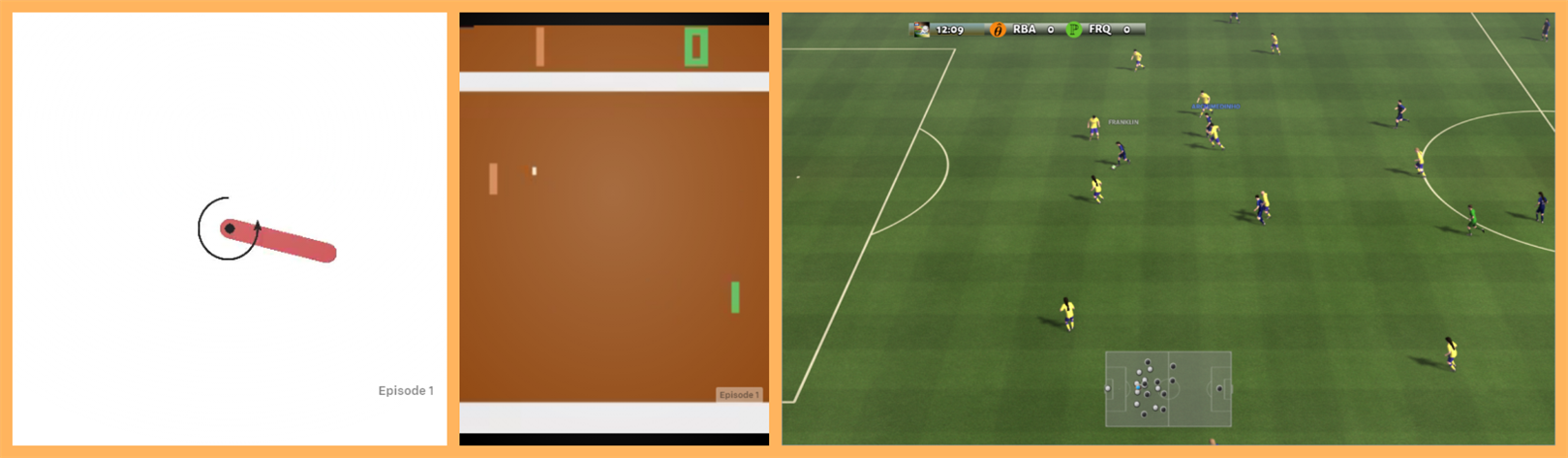}\hfill{}

\caption{\label{fig:Benchmark-games.-From}Benchmark games. From left to right:
Pendulum, Pong, Google football.}
\end{figure}

{In this section, we conduct benchmark experiments to assess the performance of Lamarckian on three benchmark games as shown in \figurename~\ref{fig:Benchmark-games.-From} in \sectionname~\ref{subsec:Sampling-Efficiency} to \sectionname~\ref{subsec:Summary}.
}

\subsection{Experimental Design\label{subsec:Experimental-Design}}

To assess the performance of Lamarckian, we conduct benchmark experiments
against the state-of-the-art library RLlib across different cluster
scales and environments. For some of the commonly used environments
in RL research (e.g., OpenAI gym) that provide only synchronous interfaces,
we wrap them with the proposed asynchronous interface to be compatible
with Lamarckian. Note that for the inherently synchronous environments,
this modification merely affects sampling efficiency as the sampled
trajectories would remain the same for the same action sequence. To
sufficiently demonstrate the efficiency of gathering and broadcasting,
we adopt an asynchronous variant of PPO built on top of an architecture
similar to that of IMPALA with the proposed Ray+ZeroMQ and tree-shaped
broadcasting method. All results are averaged over 5 independent runs
of the corresponding experiments.

For small-scale experiments (10 to 160 CPU cores), we train agents
to play Atari Pendulum and the image input version of Pong by PPO.
For large-scale experiments (2000 to 6000 CPU cores), we train agents
to play 1) Google football, a challenging video game based on physics
simulation with complex interactions and a sparse reward by PPO, and
2) vector-input version of Pong by using the PBT to adjust the learning
rate and loss ratio of PPO.
% After that, we use Lamarckian to generate behavior-diverse agents for
% Pong game by the implemented EMOGI algorithm. Finally, we apply Lamarckian to a game
% balancing test for an asynchronous commercial game, \emph{\href{https://www.lotr-risetowar.com/index.html}{The Lord of the Rings: Rise to War}}.

In all experiments except Pendulum, only a single GPU is used in the
learner process. The hyperparameters used in the experiments and the
model implementations are listed in \textcolor{black}{\tablename}~\ref{tab:Config}.
All experiments were carried out on a cluster where each computation
node has 40 Intel Xeon Gold 6148 CPU @2.4GHz. The learner always uses
an NVIDIA Tesla A100-40GB GPU.

\begin{table}[tbh]
\caption{\label{tab:Config}Hyperparameter setting and model implementation
for all benchmark instances. For PPO, we use the normalized advantage
estimate and no KL penalty. Intervals indicate the search ranges of
the corresponding parameter in PBT. The critic networks are built
on top of the policy networks, sharing the bottom part. LeakyReLU
activation is applied to each layer. }

\hfill{}\scriptsize%
\begin{tabular}{lccc}
\toprule 
\multirow{2}{*}{Benchmarks} & \multicolumn{3}{c}{PPO}\tabularnewline
 & Discount & Clip & Learning rate\tabularnewline
\midrule
Pendulum & 0.99 & 0.2 & 0.01\tabularnewline
Image Pong & 0.99 & 0.2 & 0.01\tabularnewline
Vector Pong & 0.99 & 0.2 & (0, 0.1{]}\tabularnewline
Gfootball & 0.993 & 0.2 & 0.0005\tabularnewline
\midrule
\midrule 
Benchmarks & Batch size & Batch reuse & Loss weight (policy, critic, entropy)\tabularnewline
\midrule
Pendulum & 8192 & 1 & 1, 0.5, 0.01\tabularnewline
Image Pong & 8192 & 1 & 1, 0.5, 0.01\tabularnewline
Vector Pong & 8192 & 1 & 1, (0, 1{]}, 0.01\tabularnewline
Gfootball & 32768 & 2 & 1, 0.5, 0.01\tabularnewline
\midrule
\midrule 
Benchmarks & \multicolumn{2}{c}{Policy models (hidden layers)} & Critic models (hidden layers)\tabularnewline
Pendulum & \multicolumn{2}{c}{FC(256, 128)} & FC(256, 128, 128, 64)\tabularnewline
Image Pong & \multicolumn{2}{c}{(ConvK4S2) $\times$ 4, FC(256)} & (ConvK4S2) $\times$ 4, FC(256)\tabularnewline
Vector Pong & \multicolumn{2}{c}{FC(256, 128)} & FC(256, 128)\tabularnewline
Gfootball & \multicolumn{2}{c}{FC(512, 256, 128)} & FC(512, 256, 128, 128, 64)\tabularnewline
\bottomrule
\end{tabular}\hfill{}
\end{table}

\begin{figure}[htbp]
\hfill{}\subfloat[\label{fig:pendulum}Pendulum-CPU]{\includegraphics[scale=0.45]{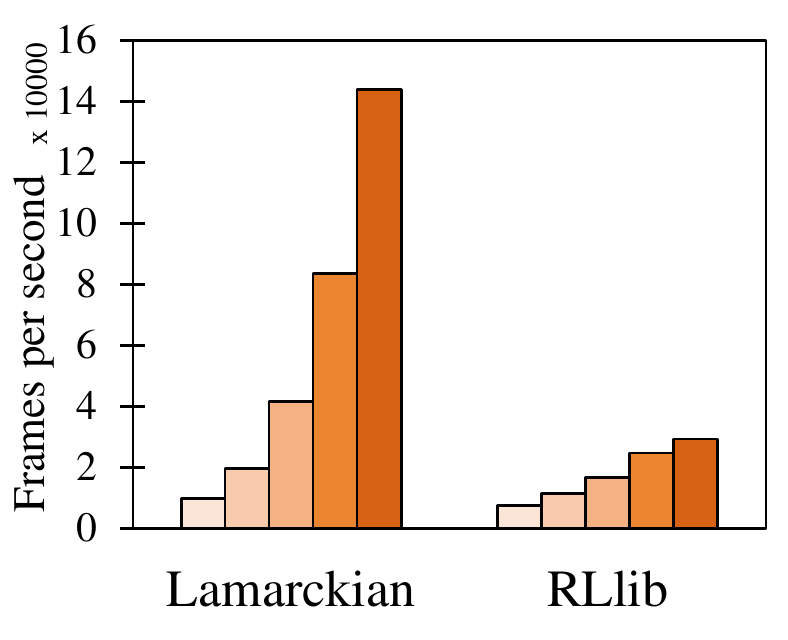}

}\hfill{}\subfloat[\label{fig:Pong}Pong-GPU]{\includegraphics[scale=0.45]{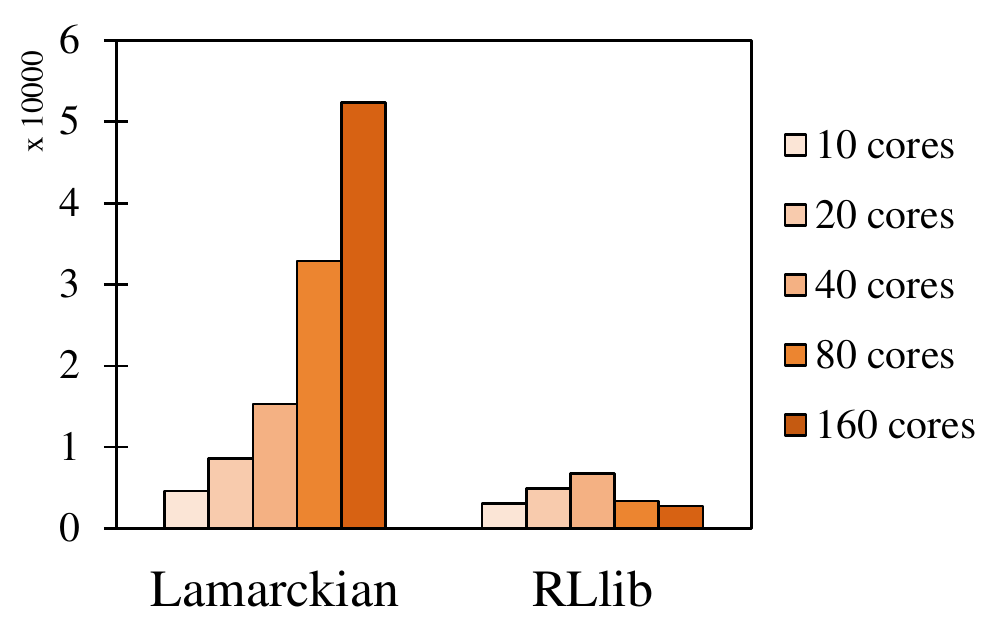}

}\hfill{}

\hfill{}\subfloat[\label{fig:Gfootball-fps}Gfootball-GPU]{\includegraphics[scale=0.45]{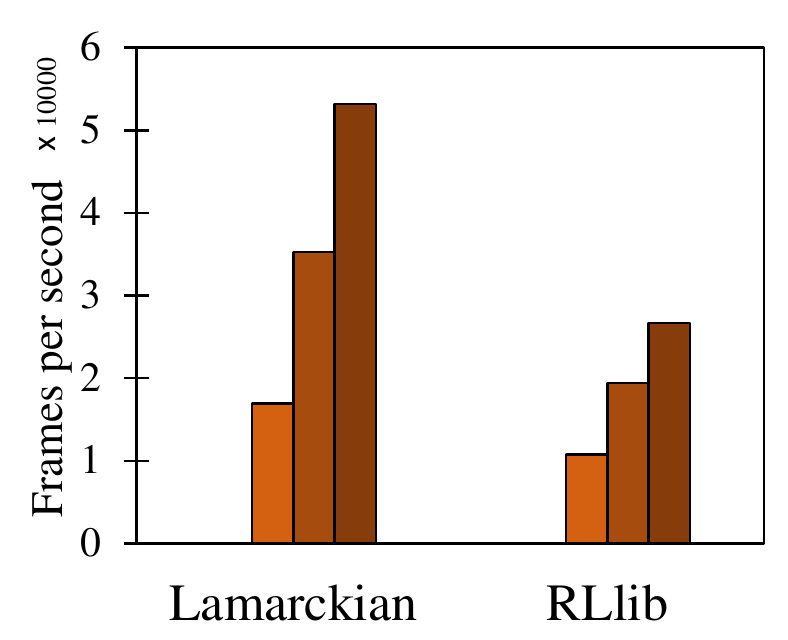}

}\hfill{}\subfloat[\label{fig:staleness}Gfootball-GPU]{\includegraphics[scale=0.45]{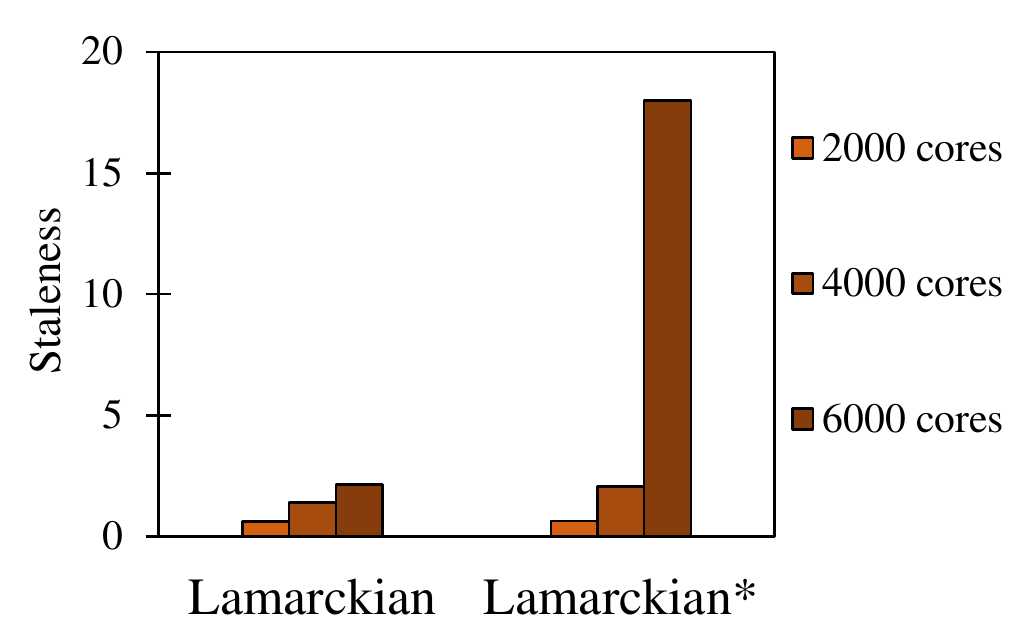}

}\hfill{}

\caption{\label{fig:Sample-efficiency-Pendulum-Pong}Sampling efficiency of
Lamarckian scales nearly linearly both on small-scale (10 to 160 CPU
cores) and large-scale (2000 to 6000 CPU cores) training instances,
which is only true for RLlib in (a) and (c). Besides, the staleness
metric of Lamarckian{*} in (d) sharply increases as the number of
CPUs grows.}
\end{figure}

\begin{table}[tbh]
\caption{\label{tab:CPU utilization-1}The average CPU utilization of Lamarckian
and RLlib when running PPO on Image Pong.}

\hfill{}\scriptsize%
\begin{tabular}{lccccc}
\toprule 
\multirow{2}{*}{Platforms} & \multicolumn{5}{c}{CPU cores}\tabularnewline
 & 10 & 20 & 40 & 80 & 160\tabularnewline
\midrule
Lamarckian & 32.2\% & 62.1\% & 96.8\% & 97.2\% & 69.4\%\tabularnewline
RLlib & 10.7\% & 21.2\% & 23.9\% & 7.5\% & 6.5\%\tabularnewline
\bottomrule
\end{tabular}\hfill{}
\end{table}

\subsection{Sampling Efficiency\label{subsec:Sampling-Efficiency}}

Sampling efficiency is measured by the average number of frames consumed
per second (fps) during the training and evaluation processes. Higher
sampling efficiency leads to a faster training speed of agents generally.
As shown in \figurename~\ref{fig:Sample-efficiency-Pendulum-Pong},
Lamarckian has near-linear scalability in all cases, which is only
true for RLlib in \figurename~\ref{fig:pendulum} and \figurename~\ref{fig:Gfootball-fps}.
In detail, on the Pendulum game, Lamarckian achieves 10k samples per second on 10 cores. Compared with the sampling efficiency with the minimum number of cores, Lamarckian reaches about 2, 4, 8, and 15 times sampling efficiency on 20, 40, 80, and 160 cores, respectively. On the Pong game, Lamarckian achieves similar acceleration on these cores. Moreover, on the complex Gfootball game, Lamarckian obtains about 17k samples per second on 2000 cores, which becomes more advantageous as the number of cores increases, reaching over 53k samples per second on 6000 cores. The results show that even in the case communicating on thousands of cores, Lamarckian can still achieve near-linear scalability.
In constrast, RLlib shows generally lower performance and degenerated in the experiment on Image Pong,
suspiciously due to its complex remote object managing mechanism and
limited communication efficiency. 

The average CPU utilization of Lamarckian is much higher than that
of RLlib when running PPO on Image Pong in \tablename~\ref{tab:CPU utilization-1}.
{Specially, Lamarckian can achieve CPU utilization
of 96.8\% and 97.2\% with one machine and two machines, respectively.
However, the CPU utilization of both platforms decreases with an increasing
number of machines, further indicating the potential improvement brought by Lamarckian.}

\subsection{Data Broadcasting Efficiency\label{subsec:Data-Broadcasting-Efficiency}}

Sample staleness, defined to measure the degree of policy-lag, is
calculated as the version difference between the policy used to draw
the samples and the policy being optimized. Note that RLlib's implementation
of PPO employs synchronous sampling (the learner stalls to wait for
new samples), thus always having a reference staleness of one. Fractional
values are possible since PPO takes more than one gradient step on
each batch of data. In an asynchronous sampling scheme, the staleness
is mainly determined by the latencies of broadcasting updated policies
and gathering samples. 

To verify the efficiency of the proposed tree-shaped data broadcasting,
we conduct ablation experiments by comparing it against asynchronous
broadcasting with a flat layout as discussed in \sectionname~\ref{subsec:Asynchronous-Tree-shaped-Data},
termed Lamarckian{*}. Moreover, we study how the staleness influences
the agents' learning efficiency, as measured by the performance achieved
given the same amount of samples.

As shown in \figurename~\ref{fig:staleness}, the staleness of Lamarckian{*}
sharply increases as the number of CPUs grows, which matches our expectation
on its scaling behavior. And the out-of-date samples also lead to
worse performance, which can be observed from \figurename~\ref{fig:PPO on Gfootball}.
Besides, although the staleness of Lamarckian is larger when on 4000
cores or more, the performance of trained agents is consistently better.
Therefore, we can conclude that the proposed tree-shaped broadcasting
can effectively reduce the side effects brought by the policy-lag
in asynchronous learning schemes.

\subsection{Performance and Training Speed\label{subsec:Performance-and-Training}}

\begin{figure*}[htpb]
\hfill{}\subfloat[\label{fig:1800cores-1}2000 cores]{\includegraphics[scale=0.16]{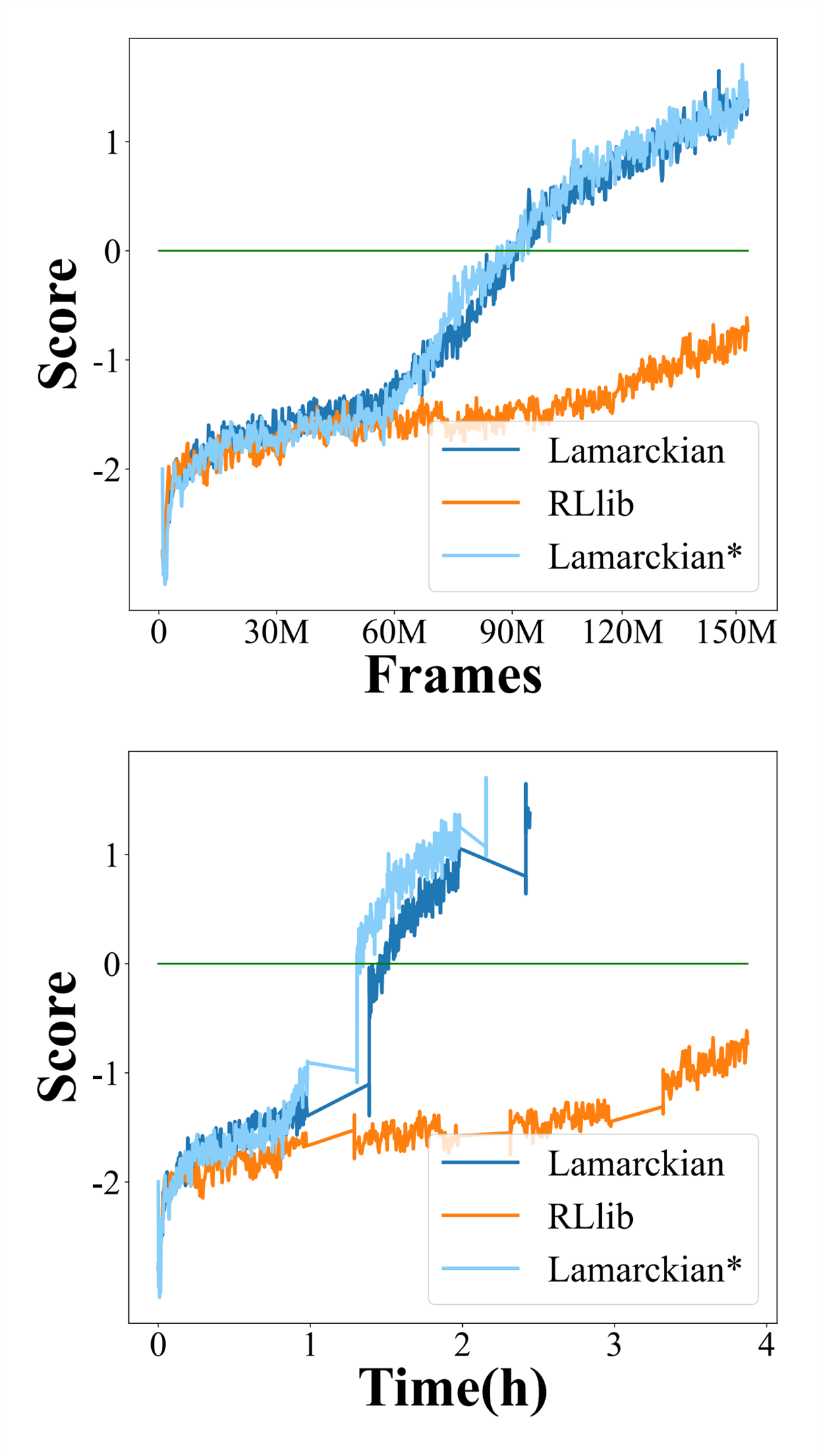}

}\hfill{}\subfloat[\label{fig:3600cores-1}4000 cores]{\includegraphics[scale=0.16]{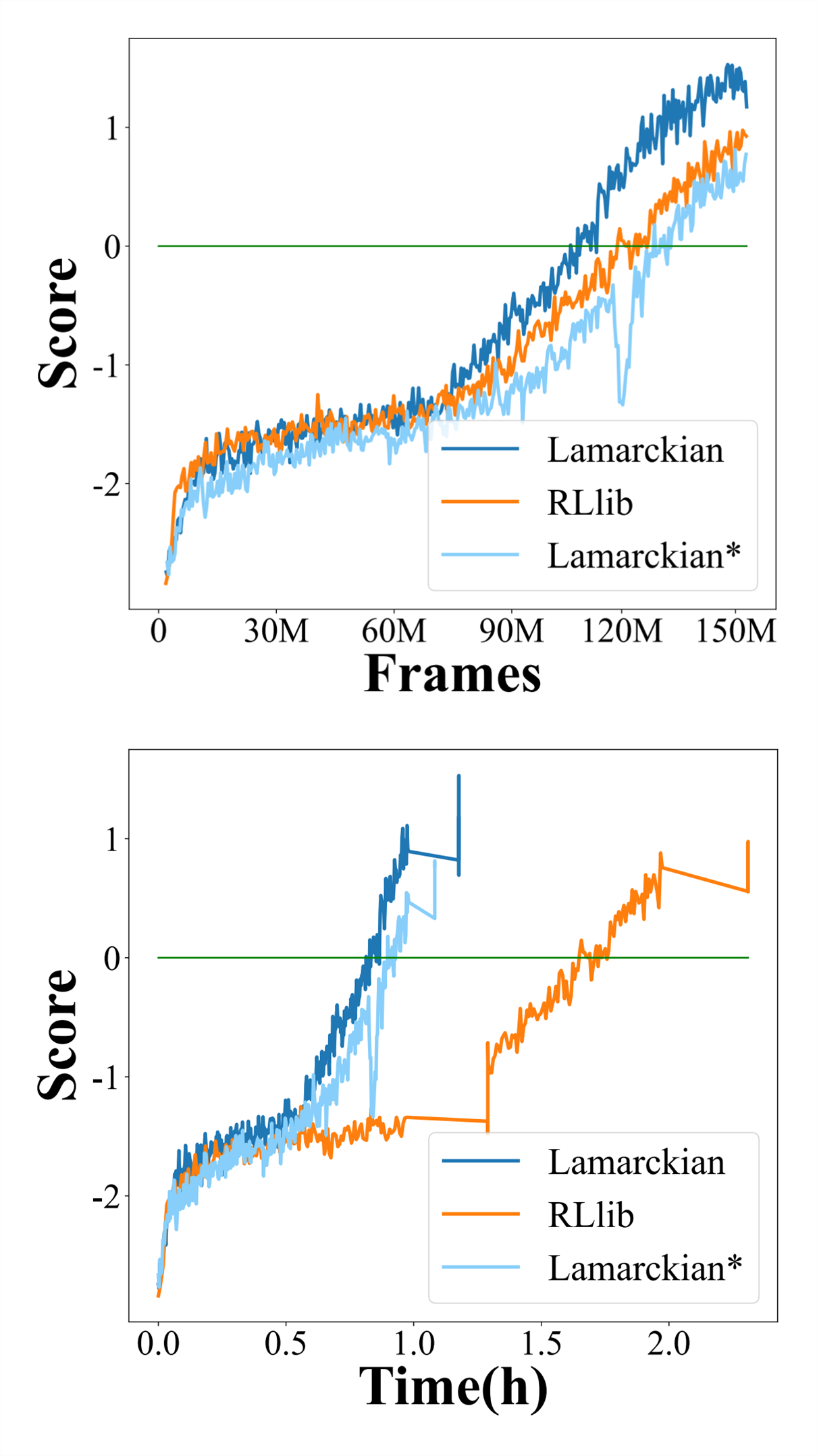}

}\hfill{}\subfloat[\label{fig:5400cores-1}6000 cores]{\includegraphics[scale=0.16]{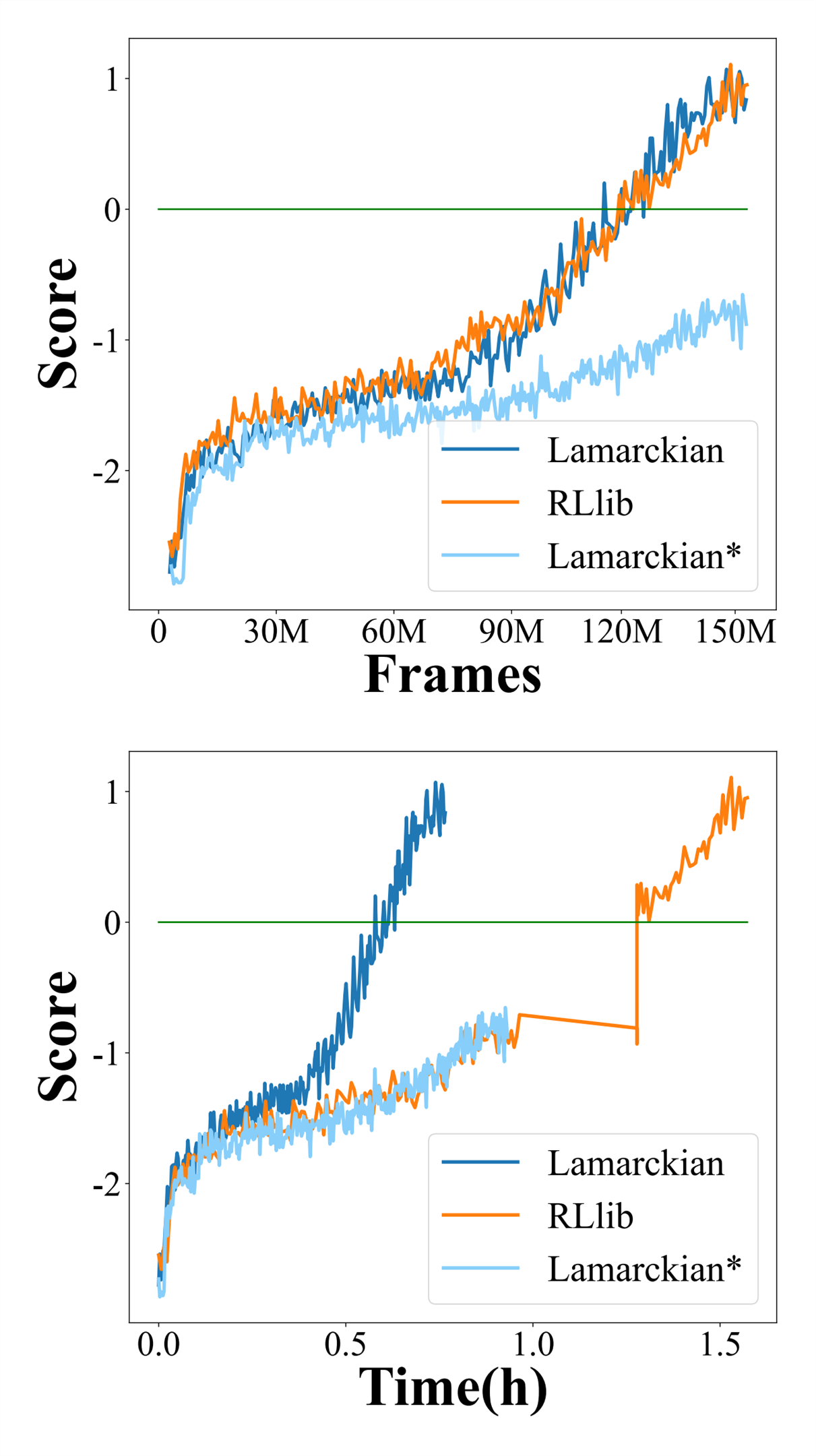}

}\hfill{}\caption{\label{fig:PPO on Gfootball}The agents' performance and the training
speed of Lamarckian, RLlib and Lamarckian{*} by the learning curves
in 150M environment frames when running PPO on Gfootball.}
\end{figure*}

\begin{figure*}[htbp]
\hfill{}\subfloat[\label{fig:1800cores-1-1}2000 cores]{\includegraphics[scale=0.16]{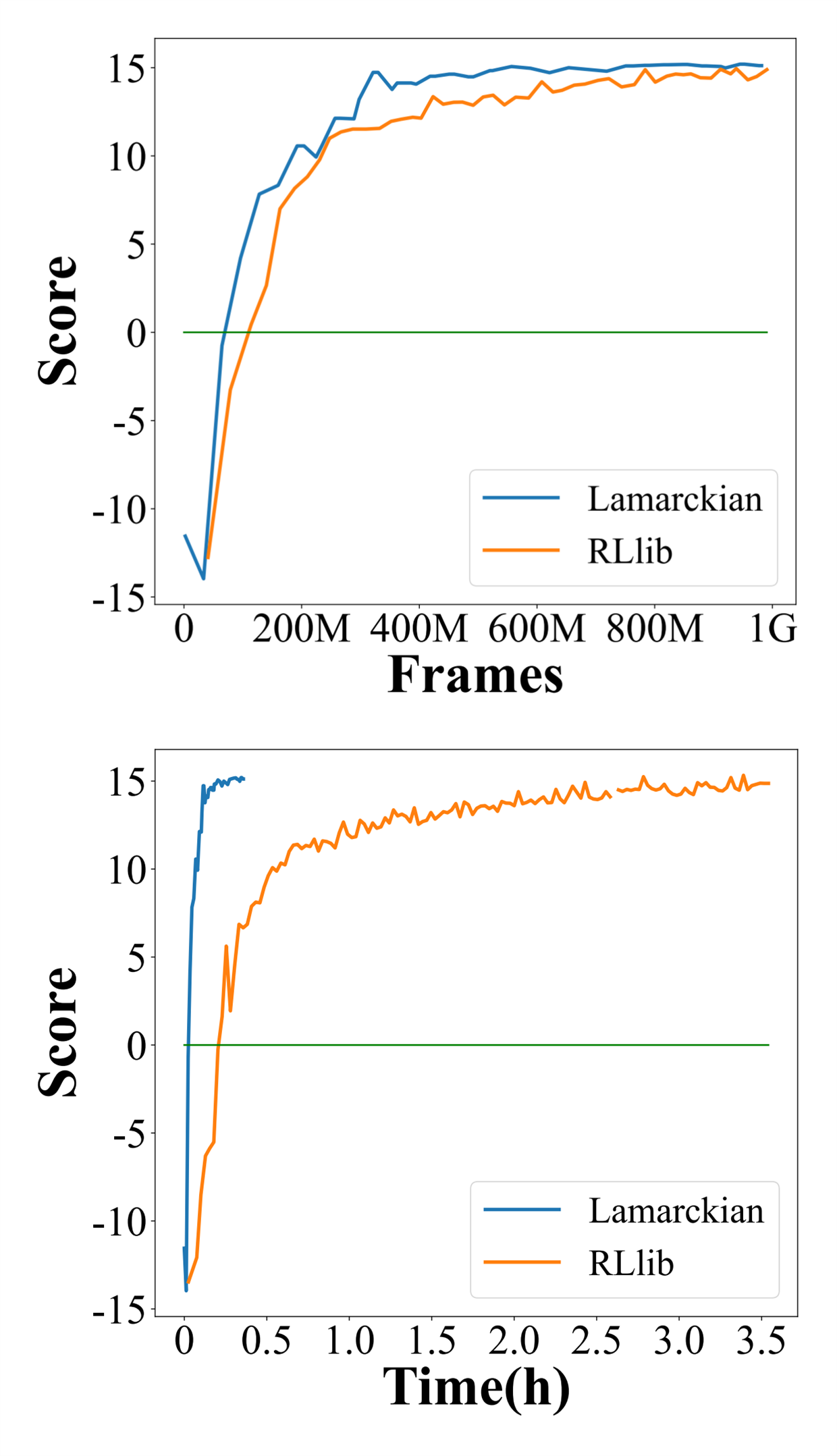}

}\hfill{}\subfloat[\label{fig:3600cores-1-1}4000 cores]{\includegraphics[scale=0.16]{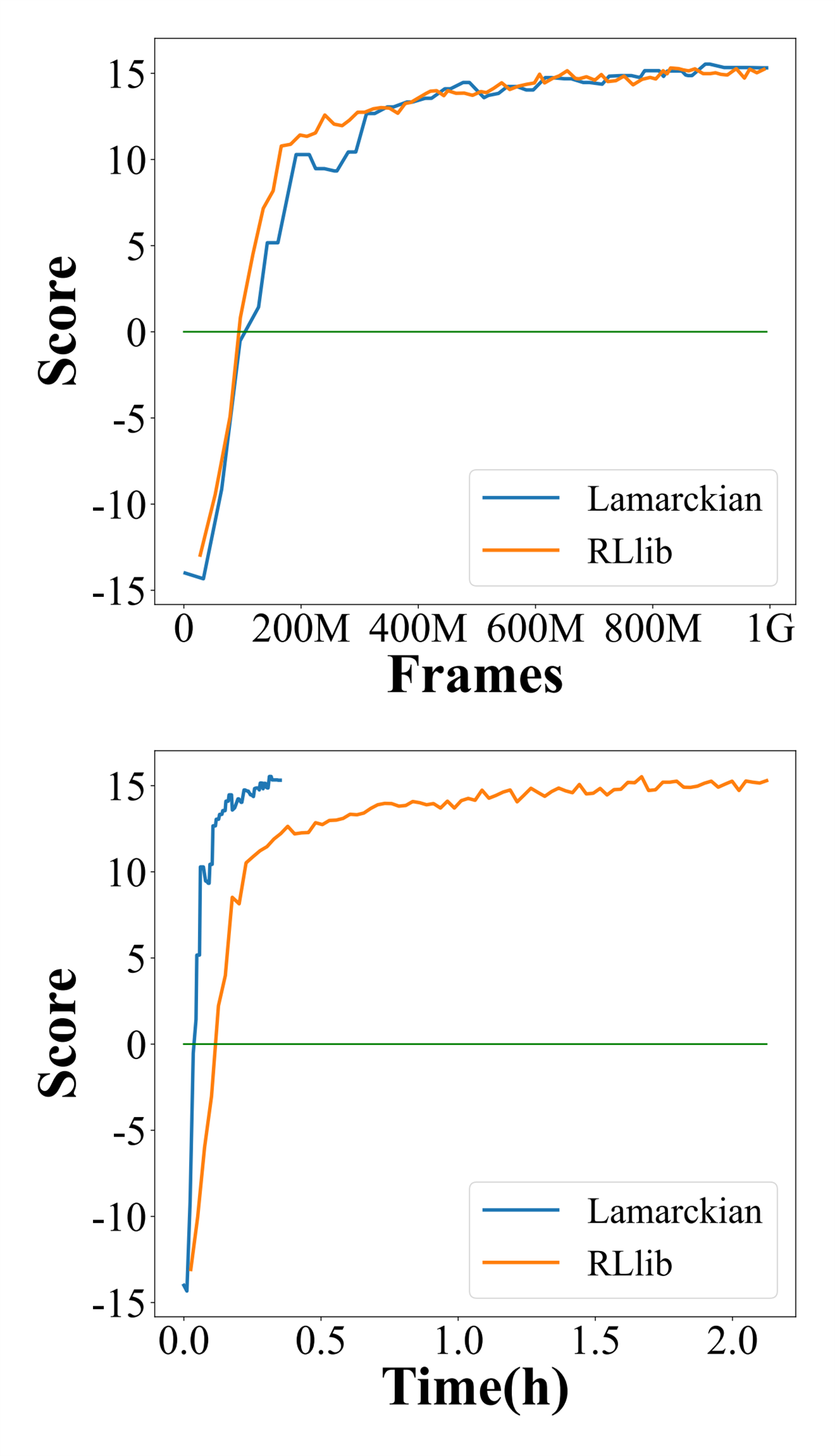}

}\hfill{}\subfloat[\label{fig:5400cores-1-1}6000 cores]{\includegraphics[scale=0.16]{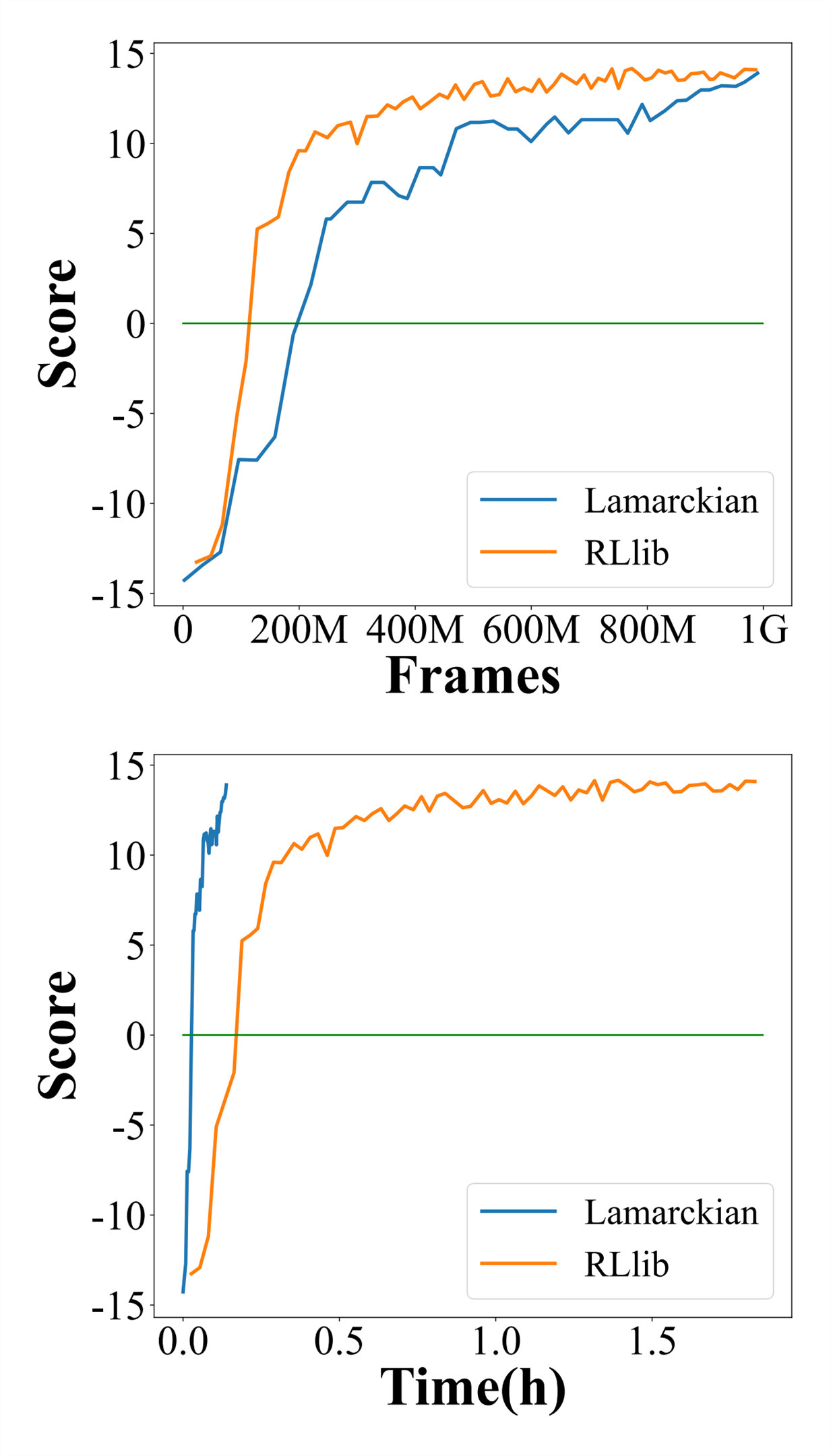}

}\hfill{}

\caption{\label{fig:PBT+PPO on Pong}The agents' performance and the training
speed of Lamarckian and RLlib by the learning curves in 1G environment
frames when running PBT+PPO on Pong.}
\end{figure*}

We investigate the agents' performance and the training speed of Lamarckian
and RLlib by the learning curves in fixed numbers of environment frames
(i.e., 150M frames for PPO on Gfootball, and 1G frames for PBT+PPO
on Pong) on the three large-scale instances. \figurename~\ref{fig:PPO on Gfootball}
and \figurename~\ref{fig:PBT+PPO on Pong} provide the learning
curves from two perspectives, with total frames and time as the horizontal
axes respectively. The two subfigures in each column represent the
same instance of CPU cores.

As shown in the two figures, Lamarckian achieves competitive scores
on both PPO for Gfootball and PBT+PPO for Pong. Moreover, the training
speed of Lamarckian is much faster than that of RLlib in both games:
i) in \figurename~\ref{fig:PPO on Gfootball}, with a larger number
of CPU cores, Lamarckian achieves as twice fast training speed as
RLlib on 6000 cores; ii) in \figurename~\ref{fig:PBT+PPO on Pong},
Lamarckian is 13 times faster than RLlib on 6000 cores.

However, when consuming the same frames, RLlib converges
faster than Lamarckian on Pong with 4000 and 6000 cores in the first
row of \figurename~\ref{fig:PBT+PPO on Pong}, but the observation
is not true in \figurename~\ref{fig:PPO on Gfootball}. The different
observations of the performance and training speed from the two figures
are highly related to the difficulties of two tasks, i.e., tasks of
different difficulties have different sensitivities to the abundance
of samples. In the experiments, the task of Gfootball, with sparse
rewards, is a much more challenging RL task than the task of Pong.
Consequently, Lamarckian achieved significantly faster convergence
(as well as higher scores) than RLlib on Gfootball due to the abundant
samples generated by the asynchronous sampling, while the limited
samples generated by the synchronous sampling in RLlib seem good enough
for the task of Pong.
Besides, as shown in \figurename~\ref{fig:PPO on Gfootball},  RLlib stagnates on 2000 cores while converging better on 4000 and 6000 cores reflects. It can be attributed to the fact that the inefficient and unstable communication mechanism of RLlib can lead to its unstable performance on large-scale CPU cores.

\subsection{{Discussion\label{subsec:Summary}}}

{
In asynchronous RL, higher sampling efficiency does
not necessarily lead to faster convergence or higher performance in
frames, but staleness plays the key role. In Lamarckian,
the staleness of samples increases as the transfer of samples takes
longer; by contrast, RLlib adopts a synchronous learning mechanism
where the staleness is fixed no matter how long it takes to transfer
the samples. Hence, despite that Lamarckian has significantly higher
sampling efficiency, its staleness also becomes larger as the number
of machines increases. Nonetheless, the proposed tree-shaped broadcasting
is able to effectively reduce the side effects brought by the policy-lag
and the communication bottleneck of learners in the asynchronous learning
procedure, thus leading to competitive performance (i.e., scores)
with a faster training speed. Furthermore, the distributed EvoRL workflow
can also accelerate the training speed for EvoRL methods. In common
practice, staleness is often fixed by synchronous sampling, while
maintaining stable staleness is particularly crucial and challenging
in asynchronous RL.}

\section{Use Cases}\label{sec:Use Cases}
In this section, we provide two use cases of Lamarckian: first, we provide a use case of how to generate behavior-diverse game AI by implementing a state-of-the-art algorithm in Lamarckian; then, we provide a use case of how Lamarckian is applied to game balancing tests for an asynchronous
RTS game.

\subsection{Generating Behavior-Diverse Game AI\label{subsec:Generating-Behavior-Diverse-Game}}

In commercial games, \emph{game AI} entertains users via human-like interactions.
For high-quality game AI, \emph{diversity of behaviors} is among the most important criteria to meet.
However, generating behavior-diverse game AI often involves rich domain knowledge and exhaustive human labors.
For example, the behavior tree \cite{Millington2018} is a rule-based method, which requires abundant expert knowledge and labor costs in designing rules. In contrast, the EC-based methods can generate strong game AI beyond human common sense by requiring little prior human knowledge   \cite{Mouret2015,Agapitos2008,Shen2020a}.

The EMOGI \cite{Shen2020a} is a recently proposed EvoRL
approach for generating diverse behaviors for game AI with little prior knowledge. 
The basic idea of EMOGI is to guide the AI agent to learn towards desired behaviors automatically by tailoring a reward function with multiple objectives, where a multi-objective optimization algorithm is adopted to obtain policies trading-off between the multiple objectives.

We implement EMOGI in Lamarckian and show a use case
on the Atari Pong game.
Specifically, apart from maximizing the win-rate/score, a Pong
agent is designed by considering \emph{active} or \emph{lazy} behavior styles according to its willingness
to make movements during a game epoch. 
The two styles are formulated
as a reward involving two objectives:

\begin{equation}
\textbf{r}(s,{a})= [f_{1}(s,{a}),f_{2}(s,{a})]^{T}, \\
\label{Pong AI}
\end{equation}
with
\begin{equation}
\left \{
\begin{array}{c}
 f_{1}(s,{a})  = \mathrm{\textrm{win}}(s) + \textrm{activeness}({a}) \\
 f_{2}(s,{a})  = \mathrm{\textrm{win}}(s) + \textrm{laziness}({a})\\
\end{array},
\right.
\end{equation}
and
\begin{equation}
\left \{
\begin{array}{c}
\textrm{activeness}({a})= w_1 \cdot \textrm{move}(a) \\
\textrm{laziness}({a})  = w_2  \cdot (1 - \textrm{move}(a))\\
\end{array},
\right.
\end{equation}
where $f_{1}(s,a)$ and $f_{2}(s,a)$ are the \emph{activeness}-related objective and the \emph{laziness}-related objective respectively;
%where $\textrm{move-rate}(a)=1-\textrm{non-move-rate}(a)$; 
$\text{win}(s)$ returns $1$ iff $s$ is a winning state, otherwise $0$;
$\textrm{move}(a)$ returns $\frac{1}{T}$ iff there is any movement in action $a$, otherwise $0$, with $T$ denoting the total number of actions in an epoch;
$w_{1}$ and $w_{2}$ are weight parameters set by users. 
Intuitively, the reward as formulated above encourages an agent to either play in an
active manner or, conversely, try to win the game with a minimal number
of movements.

\begin{figure}[htpb]
\hfill{}\includegraphics[scale=0.50]{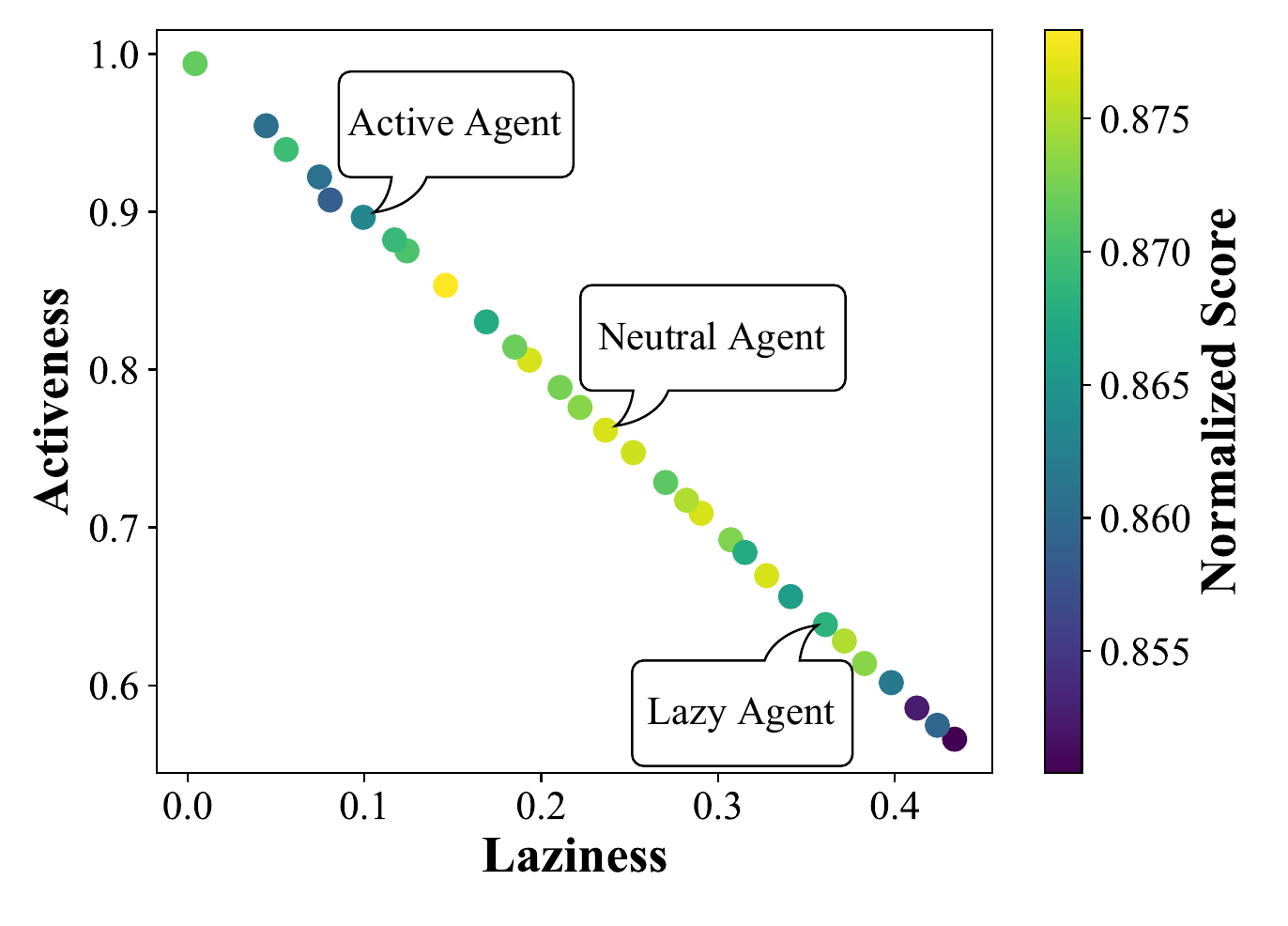}\hfill{}

\caption{\label{fig:EMOGI} The Pareto front of activeness vs. laziness values of the AI agents obtained by EMOGI using Lamarckian on Atari Pong game. Colors indicate the average normalized scores achieved by each agent.}
\end{figure}

\begin{figure*}[htbp]
\hfill{}\subfloat[\label{fig:arena} Fighting arena]{\includegraphics[scale=0.3]{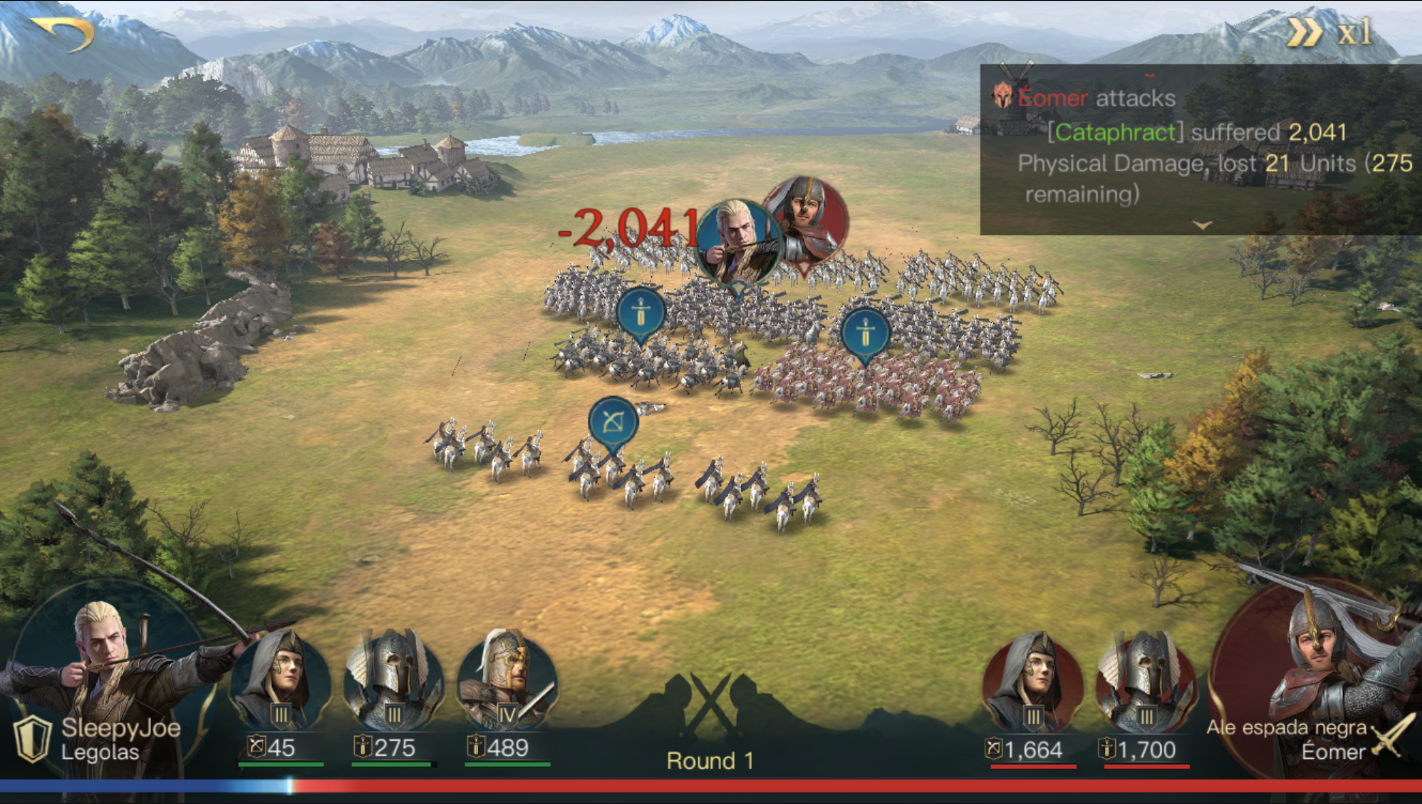}

}\hfill{}\subfloat[\label{fig:Pareto}Final solution set]{\includegraphics[scale=0.48]{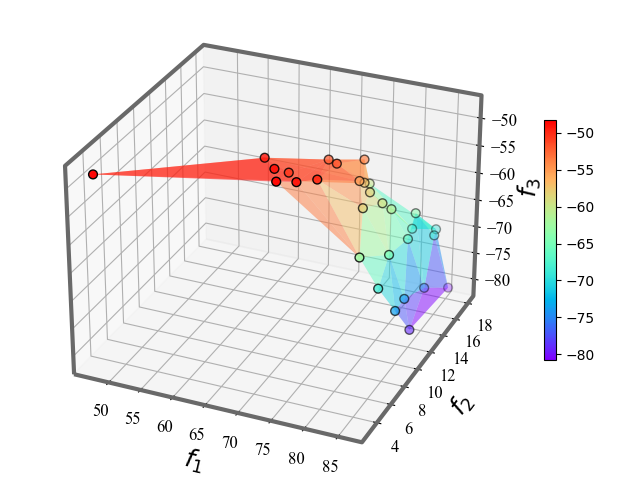}

}\hfill{}

\caption{\label{fig:The-optimal-solutions}(a) is a typical hero-led fighting
arena of The Lord of the Rings: Rise to War. (b) is the final solution
set obtained by Lamarckian to a three-objective optimization problem
in the game balancing test. $f_{1}$ is the battle damage difference,
$f_{2}$ is the remaining economic resources multiples, and $f_{3}$
is the strength of the weakest team.}
\end{figure*}

As evidenced in~\figurename~\ref{fig:EMOGI}, the AI agents formed different play styles in terms of activeness/laziness, while achieving similar normalized scores (defined as $\frac{score_{\text{self}}-score_{\text{enemy}}+score_{\max}}{2score_{\max}}\in[0, 1]$). 
On the one hand, the agents with relatively lower competitiveness (i.e. those on the two corners) have survived due to their distinguished behavior styles -- either too active or too lazy.
On the other hand, the agents with
relatively higher competitiveness ((i.e. those in the center) have also survived despite that the behavior styles are neutral.
%During the process of EvoRL, when two agents have similar high win-rates, the distinguished behavior styles enable both of them to survive in the population. Besides, it is also evidence that (the central agents in the figure). 
In practice, such wide spectrum of behavior-diverse game AI would enable richer user experience.

\subsection{Game Balancing in RTS Game\label{subsec:Application-to-An}}

%{This section provides a case study of Lamarckian to an asynchronous commercial game in real-world scenarios. Since this case study is pre-market testing, we only present the methodology and solutions without comparative experiments.} 

In commercial games, user satisfaction is always influenced by game
balancing \cite{MesentierSilva2019}. In an imbalanced game, there
maybe exist an unbeatable strategy or a role getting the player frustrated.
The game designers aim to develop a balanced game by weakening the
most imbalanced strategies or roles obtained by the game balancing
test. Traditionally, the enumeration methods or reinforcement methods have been adopted to the problem in industry. However, they are low-efficient since only one solution is obtained in each run. In contrast, the EC-based methods are more efficient by obtaining a set of diverse solutions in one run.

From the optimization point of view, the game balancing test
is a non-differentiable optimization problem that involves multiple
objectives to be considered simultaneously.
We apply Lamarckian to a three-objective optimization problem in the
game balancing test for the asynchronous commercial RTS game, The
Lord of the Rings: Rise to War, aiming to obtain the most imbalanced
and strongest lineups. First, we encode a lineup as the decision variables
of a candidate solution, including heroes (51 types) and soldiers
(two to three types); each hero is armed with four types of skills
and 79 types of equipment. Thus, the search space is up to 50,000
units. A typical hero-led fighting arena is shown \figurename~\ref{fig:arena}.
Second, we formulate the game balancing test problem as a multi-objective
combinatorial optimization (maximization) problem associated with
three objectives: i) the battle damage difference ($f_{1}$) calculated
by the remaining strength difference between the strongest team and
the weakest team in a lineup, ii) the remaining economic resources
multiples ($f_{2}$) calculated by the remaining economic resources
of the strongest team divided by those of the weakest team, and iii)
the strength of the weakest team ($f_{3}$) to improve the overall
strength of a lineup. Third, we use crossover and mutation operators
of discrete variables similar to those in the classic Traveling Salesman
Problem to ensure the validity of offspring and the non-repetitiveness
of teams. Then, to evaluate a candidate solution, all teams in the
solution should play against each other to obtain the first two objectives,
and then battle with the baseline lineups to determine the strength
of the weakest team. Finally, we adopt NSGA-II, a classic EC algorithm
for multi-objective optimization, as the optimizer to iteratively
perform non-dominated sorting on a population of candidate solutions
evolving towards the Pareto front. The final population is an approximation
to the Pareto optimal set. {The solutions in the Pareto
optimal set mean that there is no solution better than the other solutions
on all the objectives and they are Pareto non-dominated.}

As shown in \figurename~\ref{fig:Pareto}, a solution with a remaining
economic resources multiple larger than two (i.e., $f_{2}>2$) is
considered \emph{imbalanced}. Apart from the solution set itself,
we also obtain some interesting observations. For example, the outlier
solution at the left corner indicates the strong conflicts between
$f_{1}$ and $f_{3}$, and $f_{2}$ and $f_{3}$, respectively. It
means that a set of overall very strong lineup leads to less battle
damage differences and remaining economic resources multiples. {Eventually,
the game designers will select the solutions with overall strong lineups
(i.e., the red points) and modify the game parameters to weaken these
lineups. The above testing and modification processes will loop until
the game is balanced. The game balancing test of each loop only takes
about ten minutes by a tailored game simulator, which is acceptable
in the industry.}

\section{Conclusion}\label{sec:Conclusion}

This paper introduces Lamarckian \textendash{} an open-source high-performance
scalable platform tailored for evolutionary reinforcement learning.
To meet the requirements of applications in large-scale distributed
computing environments (e.g. the asynchronous commercial game environments),
Lamarckian adopts a tree-shaped data broadcasting method as well as
the asynchronous Coroutine-based MDP interface. To accelerate the
training of agents, Lamarckian couples Ray with ZeroMQ to take advantage
of both. From the software engineering perspective, Lamarckian also
provides good flexibility and extensibility by well decoupled objective-oriented
designs. The performance of Lamarckian has been evaluated on large-scale
benchmark tests with up to 6000 CPU cores, in comparison with the
state-of-the-art RLlib. 
Additionally, we provide two use cases of Lamarckian.
In the first use case, we apply Lamarckian to
generating behavior-diverse game AI by implementing a recently proposed
EvoRL algorithm. 
In the second use case, we apply Lamarckian to multi-objective game balancing test for an asynchronous
commercial real-time strategy game.

In order to match the asynchronous MDP interface in Lamarckian, users need to transfer the original synchronous MDP interfaces into asynchronous ones by following unified workflow. 
However, it is worthy of such additional implementation labor as it brings substantial performance improvement. Further, the ready-to-use
modules in Larmackian also brings extra benefits to users.

In summary, Lamarckian is a high-performance, easy-to-use,
and scalable platform for researchers and engineers to take instant
adventures.

% \section*{Acknowledgment}
% This work was supported by the National Natural Science Foundation
% of China (No. 61906081), the Shenzhen Science and Technology Program
% (No. RCBS20200714114817264), the Guangdong Provincial Key Laboratory
% (No. 2020B121201001), and the Program for Guangdong Introducing Innovative
% and Entrepreneurial Teams (Grant No. 2017ZT07X386).

\bibliographystyle{IEEEtran}
\bibliography{main}

\begin{IEEEbiography}[{\includegraphics[width=1in,height=1.25in,clip,keepaspectratio]{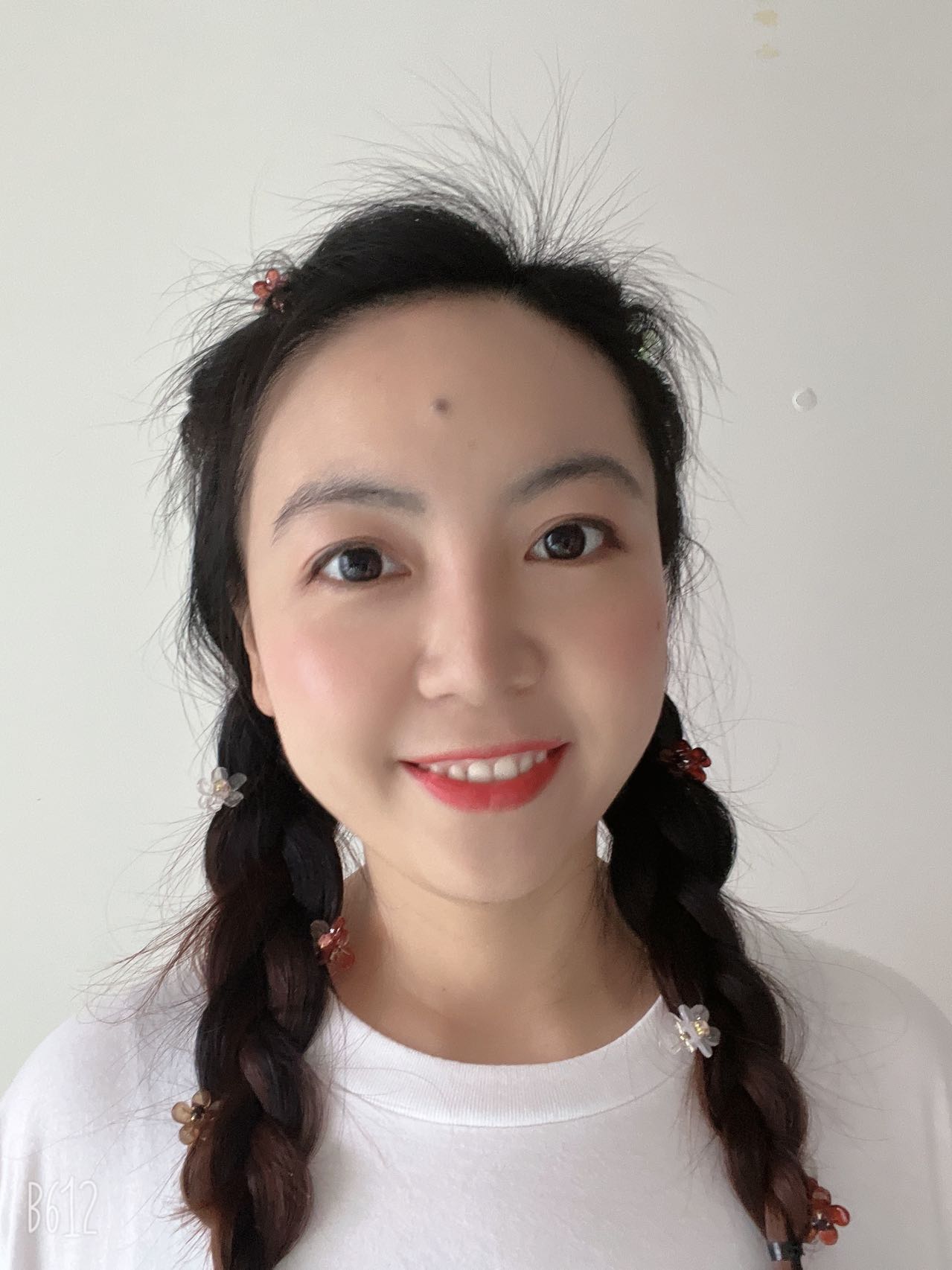}}]{Hui Bai}
 received the B.Sc. and M.Sc. degrees in software engineering from
Xiangtan University, Xiangtan, China, in 2014 and 2017 respectively. She is currently pursuing
the Ph.D. degree with the Department of Computer Science and Engineering,
Southern University of Science and Technology, China.

Her main research interests include evolutionary algorithms and their applications to reinforcement learning.
\end{IEEEbiography}

\begin{IEEEbiography}[{\includegraphics[width=1in,height=1.25in,clip,keepaspectratio]{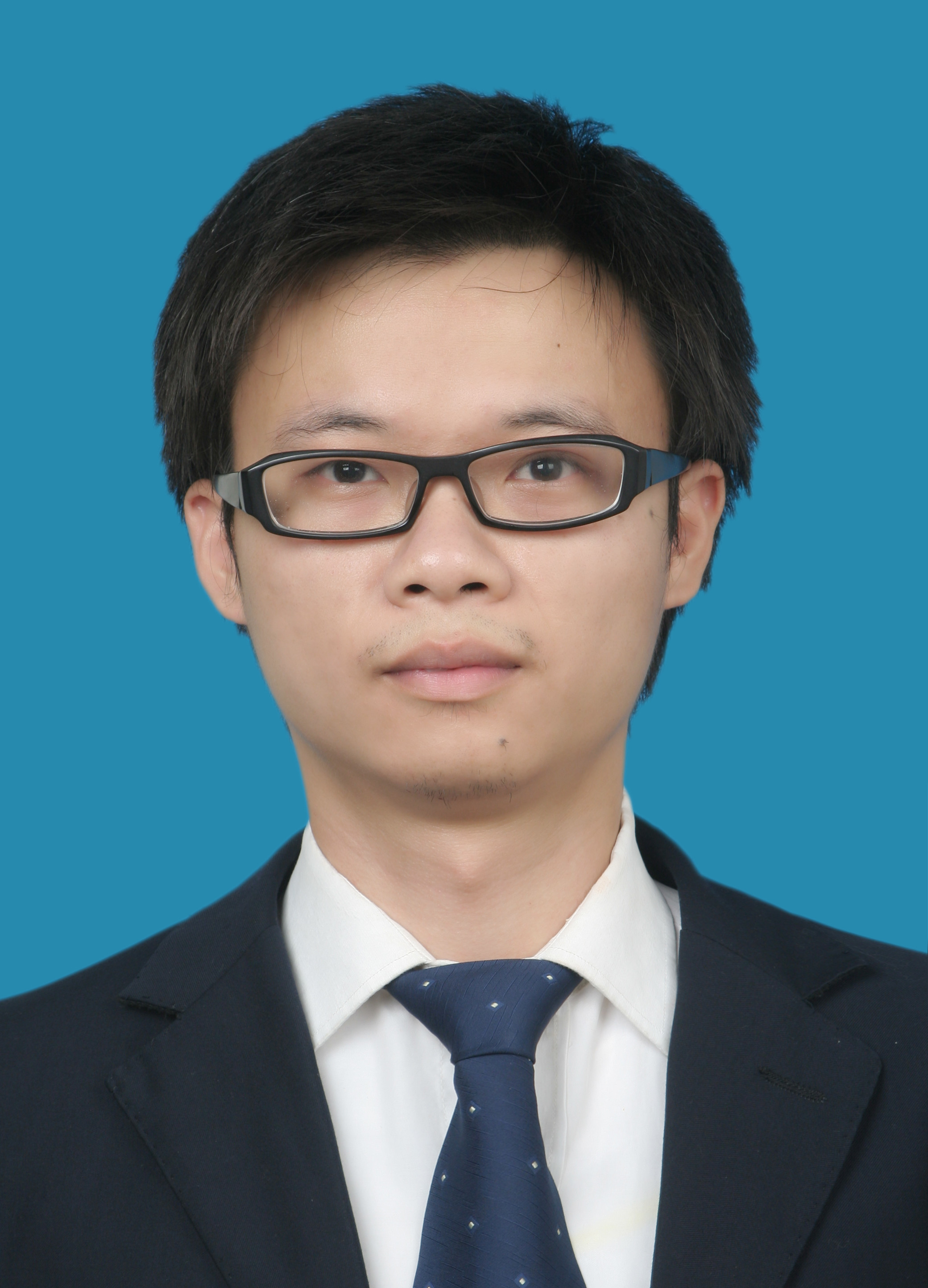}}]{Ruimin Shen}
received the Ph.D. degree in applied mathematics and M.Sc. degree in computer science from Xiangtan University of China, in 2015 and 2012, respectively. He is currently the researcher of the Game AI research team of NetEase Games AI Lab, Guangzhou, China. 

His research interests include evolutionary algorithms, reinforcement learning and their applications to online games.
\end{IEEEbiography}

\begin{IEEEbiography}[{\includegraphics[width=1in,height=1.25in,clip,keepaspectratio]{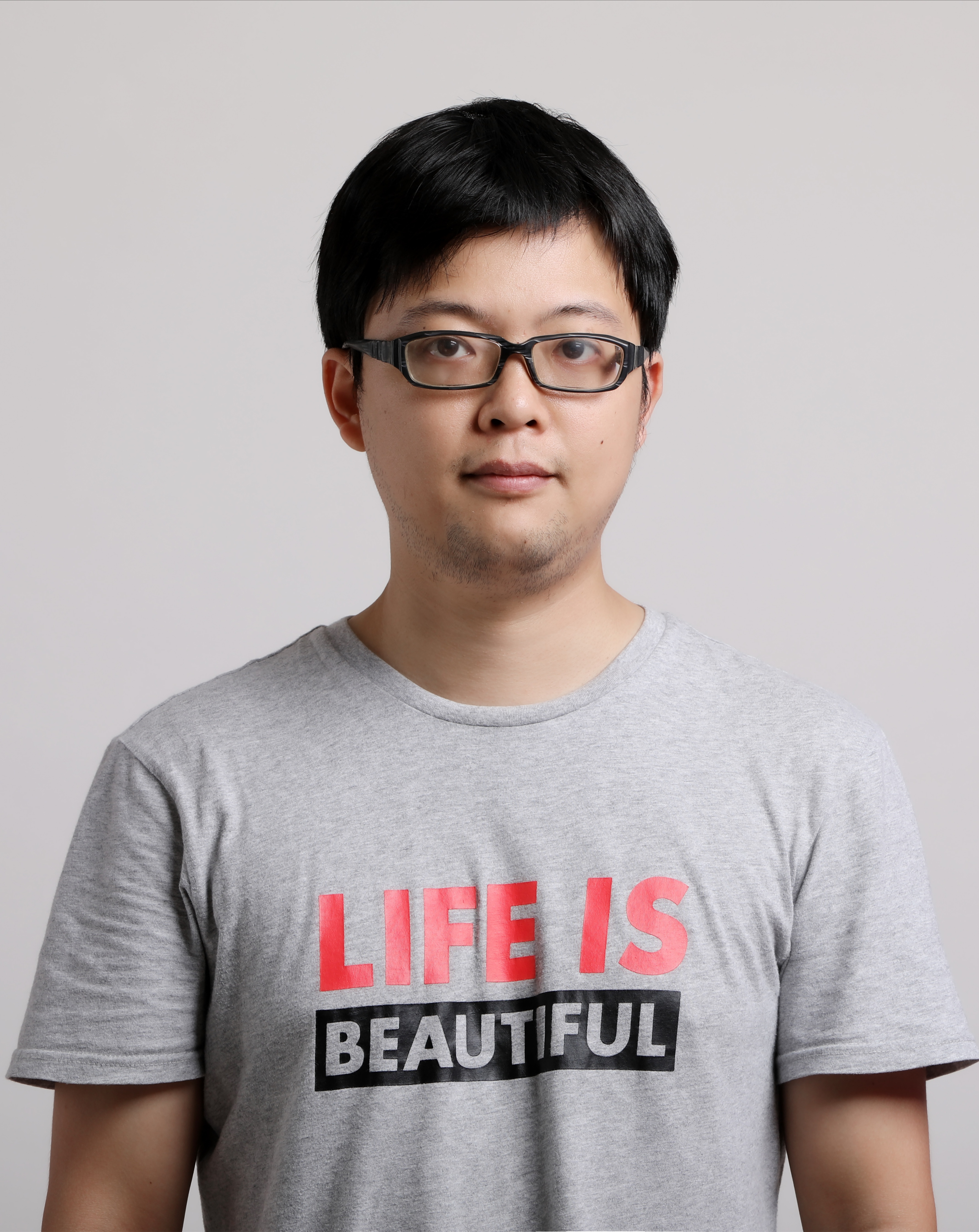}}]{Yue Lin}
received the M.Sc. degree in computer science from Zhejiang University, Hangzhou, China in 2013,  and the B.Sc. degree in control science and engineering from Zhejiang University, Hangzhou, China in 2010.  He is currently the director of NetEase Games AI Lab, Guangzhou, China.

His research interests include computer vision, data mining, reinforcement learning and their applications to online games.
\end{IEEEbiography}

\begin{IEEEbiography}[{\includegraphics[width=1in,height=1.25in,clip,keepaspectratio]{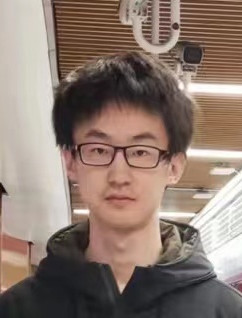}}]{Botian Xu}
 received his bachelor's degree in Computer Science from Southern University of Science and Technology. He is currently a researcher at Institution of Interdisciplinary Information Science, Tsinghua University. 
 
 His interests focus on deep reinforcement learning and its applications.
\end{IEEEbiography}

\begin{IEEEbiography}[{\includegraphics[width=1in,height=1.25in,clip,keepaspectratio]{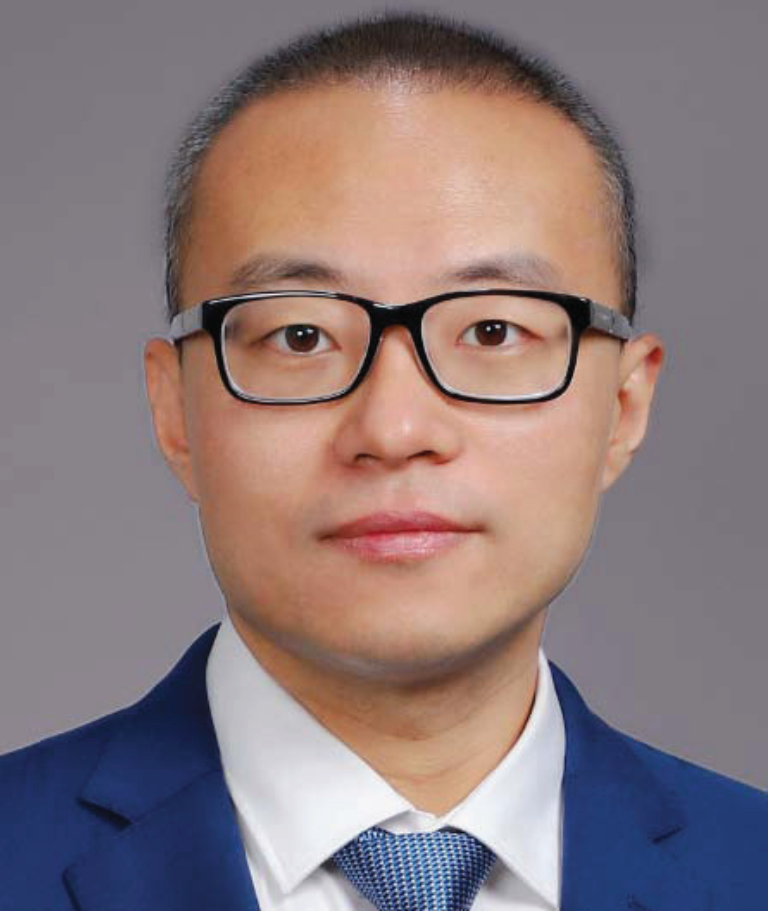}}]{Ran Cheng}
 (M'2016-SM'2021) received the B.Sc. degree from Northeastern University,
Shenyang, China, in 2010, and the Ph.D. degree from the University
of Surrey, Guildford, U.K., in 2016. He is currently an Associate
Professor with the Department of Computer Science and Engineering,
Southern University of Science and Technology, China. 

His research interests mainly fall into the interdisciplinary fields across evolutionary computation and other major AI branches such as statistical learning and deep learning, aiming to provide end-to-end solutions to optimization/modeling problems in scientific research and engineering applications.

He is a recipient of the 2019 IEEE Computational
Intelligence Society Outstanding Ph.D. Dissertation Award, the 2018 and 2021 IEEE TRANSACTIONS ON EVOLUTIONARY
COMPUTATION Outstanding Paper Awards, and the
2020 IEEE Computational Intelligence Magazine Outstanding Paper Award.
He serves as Associate Editors of the IEEE TRANSACTIONS ON EVOLUTIONARY
COMPUTATION, the IEEE TRANSACTIONS ON ARTIFICIAL INTELLIGENCE, and
the IEEE TRANSACTIONS ON COGNITIVE AND DEVELOPMENTAL SYSTEMS. He is
the Chair of IEEE Computational Intelligence Society Shenzhen Chapter.
\end{IEEEbiography}

\end{document}